\documentclass[pdflatex,sn-aps]{sn-jnl}

\usepackage[T1]{fontenc}
\usepackage{amsmath,amssymb,amsfonts}
\usepackage{graphicx}
\usepackage{booktabs}
\usepackage{textcomp}
\usepackage{xcolor}
\usepackage{tabularx}
\usepackage{url}
\usepackage{xurl}
\usepackage{hyperref}
\usepackage{enumitem}
\usepackage{float}
\usepackage{adjustbox}
\usepackage{subcaption}
\usepackage{subfiles}
\jyear{2023}%

\theoremstyle{thmstyleone}%

\theoremstyle{thmstyletwo}%

\theoremstyle{thmstylethree}%

\begin{document}

\title[Cost-Sensitive Deep Neural Networks for Brain Tumor Detection]{Explainable Cost-Sensitive Deep Neural Networks for Brain Tumor Detection from Brain MRI Images considering Data Imbalance}

\author*[]{\fnm{Md Tanvir Rouf} \sur{Shawon}}\email{shawontanvir95@gmail.com}
 \equalcont{All authors contributed equally to this work.}

\author[]{\fnm{G. M. Shahariar} \sur{Shibli}}\email{sshibli745@gmail.com}
 \equalcont{All authors contributed equally to this work.}
 
\author[]{\fnm{Farzad} \sur{Ahmed}}\email{farzadahmed6@gmail.com}
 \equalcont{All authors contributed equally to this work.}

 \author[]{\fnm{Sajib Kumar Saha} \sur{Joy}}\email{joyjft@gmail.com} 
 \equalcont{All authors contributed equally to this work.}

\affil[]{\orgdiv{Department of Computer Science and Engineering}, \orgname{Ahsanullah University of Science and Technology}, \orgaddress{\city{Dhaka}, \postcode{1208}, \country{Bangladesh}}}

\abstract{
This paper presents a research study on the use of Convolutional Neural Network (CNN), ResNet50, InceptionV3, EfficientNetB0 and NASNetMobile models to efficiently detect brain tumors in order to reduce the time required for manual review of the report and create an automated system for classifying brain tumors. An automated pipeline is proposed, which encompasses five models: CNN, ResNet50, InceptionV3, EfficientNetB0 and NASNetMobile. The performance of the proposed architecture is evaluated on a balanced dataset and found to yield an accuracy of $99.33$\% for fine-tuned InceptionV3 model. Furthermore, Explainable AI approaches are incorporated to visualize the model's latent behavior in order to understand its black box behavior. To further optimize the training process, a cost-sensitive neural network approach has been proposed in order to work with imbalanced datasets which has achieved almost $4$\% more accuracy than the conventional models used in our experiments. The cost-sensitive InceptionV3 (CS-InceptionV3) and CNN (CS-CNN) show a promising accuracy of $92.31$\% and a recall value of $1.00$ respectively on an imbalanced dataset. The proposed models have shown great potential in improving tumor detection accuracy and must be further developed for application in practical solutions. We have provided the datasets and made our implementations publicly available at - \url{https://github.com/shahariar-shibli/Explainable-Cost-Sensitive-Deep-Neural-Networks-for-Brain-Tumor-Detection-from-Brain-MRI-Images}}




\maketitle
\section{Introduction}\label{sec1}
A brain tumor is a group of abnormal cells that grows in the brain. It can strike anyone at practically any age. It may even change from one treatment session to the next, but the effects will vary from person to person. Brain tumors can arise anywhere on the brain, in various picture intensities, and in a range of shapes and sizes. There are two types of brain tumors: malignant and benign. Benign brain tumors are characterized by a homogeneous structure and the absence of malignant cells. They can be radio-logically monitored or surgically eradicated altogether, and they rarely grow back. Malignant brain tumors are composed of cancer cells and have diverse structures \cite{naik2014tumor}. Detection of tumors at an early stage is very important as the early start of treatment can cure or lessen the effect. When an MRI report is generated it needs confirmation from a doctor that it is really a tumor. If this process can be automated the extra time of confirmation from doctors can be saved. A lot of works \cite{sapra2013brain,amin2020distinctive,wu2007brain} have been done on detecting brain tumors using different machine learning algorithms and neural networks. CNN is one of the most used neural network architectures on image classification tasks. That is why CNN has been widely used \cite{ezhilarasi2018tumor,hossain2019brain} to detect brain tumor from an MRI image of a human brain. The pre-trained large models are recently popular for detecting brain tumor from MRI images. Some works \cite{khan2020brain,harish2020mri,saleh2020brain} leveraged ResNet50 and other large pre-trained models.
\par The datasets available for brain tumor detection task are mostly imbalanced. In this study, we have proposed a pipeline to deal with the data imbalance. The idea of cost-sensitive neural network is employed where the weight is biased towards the class with a comparatively low number of samples. Several works \cite{zhou2005training,li2017cost,xu2020mscs} can be referred who worked with cost-sensitive neural network for different detection tasks. In this study, we explored different CNN and CNN-based pre-trained models to detect brain tumors in MRI images. Moreover, we compared the performance of the standard and cost-sensitive deep neural networks trained on an imbalanced dataset.
\par The performance of a model on a particular task can be measured mathematically using different metrics but, we have no idea how these complex neural networks make their predictions, thus it is completely vague to us. As the effectiveness of machine learning and neural network-based models increases, so does the speed of various complex computations, regressions, and classifications. But there is a substantial danger involved. Since the models are impenetrable to us and we as users are required to place our trust in them, we rarely attempt to grasp what takes place inside of them. Without any hesitation, we also rely on the models. If we could understand how a model creates any form of judgment, that would be interesting.
Besides, a neural network may predict new data based on previously learned weights, but we are unable to see the weights or the region where the model is making the prediction. Here comes the Explainable AI which is a cutting edge technology to visualize the black box of a model. In our work, some techniques to interpret gradient based models specially trained for image classification are used. We employed two different types of Explainable AI techniques: Gradient based XAI and Perturbation based XAI. We leveraged gradient based approaches such as Grad-CAM, Grad-CAM++, Saliency Map, SmoothGrad, Score-CAM, Faster SmoothGrad and a perturbation based approach LIME to analyze the weights and the inner workings of a neural network. We have explained the output of the pre-trained models utilized in the proposed methodology using LIME, and the CNN model using gradient based explainable AI techniques. Moreover, two $2$ distinct datasets - one of which is a balanced and the other one is imbalanced have been employed to train the models in this study. In summary, we have made the following contributions in this paper:
\begin{itemize}
    \item We have proposed a pipeline where a brain tumor detection task can be performed for both the balanced and imbalanced dataset. 
    \item We have employed the concept of a cost-sensitive neural network for an imbalanced dataset and compared the performance of a standard convolutional neural network (CNN) and four pre-trained CNN models (ResNet50, InceptionV3, EfficientNetB0, and NASNetMobile). We used two separate datasets are used, one of which is balanced and the other of which is imbalanced to serve this purpose.
    \item Explainable AI techniques like Grad-CAM, Grad-CAM++, Saliency Map, SmoothGrad, Score-CAM and Faster Score-CAM are used to explain the prediction of the CNN model and LIME is used to explain the output of the pre-trained models. 
\end{itemize}

The rest of the paper is organized as follows: Some of the related earlier works are presented in section \ref{sec2}. Section \ref{sec3} describes the specifications of the datasets. In section \ref{sec4}, some key background studies regarding the cost-sensitive neural network, explainable AI techniques, and performance evaluation metrics are briefly covered. Section \ref{sec5} provides an explanation of the proposed methodology and the architecture of the models are reported in section \ref{sec6}. The experimental results are discussed in sections \ref{sec7} and \ref{sec8} respectively. Section \ref{sec9} highlights the constraints and potential future directions of our work. In section \ref{sec10}, the study concludes with a final statement.

\section{Related Works}\label{sec2}
We have included a brief summary of earlier studies that are relevant to our research in this section. The overview is roughly split into three sections: (a) studies connected to the identification of brain tumors (b) works incorporating explainable artificial intelligence and (c) works that addressed the problem of class imbalance by utilizing cost-sensitive deep neural networks.

\subsection{Brain Tumor Detection}
Early detection of brain tumors can lead to more successful treatments with fewer complications or side effects. Knowing the type and size of the tumor helps doctors determine which treatment plan is best for each patient. We broadly divided earlier works into three categories based on the need to develop systems that can analyze medical images and accurately identify areas of the brain that may have tumors or other irregularities: (a) detection systems based on traditional machine learning (b) deep learning based detection systems and (c) systems built on top of pre-trained large architectures. Table \ref{tab:reference-sum} illustrates the summary of the existing works on brain tumor detection discussed in this study.

\noindent\textbf{(a) \underline{Machine Learning}}:
The conventional machine learning techniques, such as \textit{K-Nearest Neighbor (KNN), Decision Tree (DT), Support Vector Machine (SVM)} and \textit{Adaptive Boosting (AdaBoost)} were used in plenty of studies. Naik et al. \cite{naik2014tumor} used a Decision Tree (DT) to classify brain tumors from CT scan brain images, and their model had a 96\% accuracy rate. Support Vector Machine (SVM) was employed in several studies as a classifier following feature extraction to identify brain tumors: Shil et al. \cite{shil2017improved} used Discrete Wavelet Transform (DWT) for feature extraction and Principal Component Analysis (PCA) for dimension reduction, Mathew et al. \cite{mathew2017tumor} used Wavelet Transform (WT) for feature extraction, Singh and Kaur \cite{singh2012classification} used Gray-Level Co-occurrence Matrix (GLCM) for feature extraction, and Amin et al. \cite{amin2020distinctive} attempted to classify MRI at the image and lesion levels. 
Ramteke and Monali \cite{ramteke2012automatic} extracted statistical texture feature sets from normal and abnormal photos and utilized K-Nearest Neighbor (KNN) as a classifier, achieving an accuracy of 80\%.

\noindent\textbf{(b) \underline{Deep Learning}}:
Deep learning systems are capable of recognizing patterns in scans and point up possible problem areas to further examine. As they automatically extract highly discriminative features in the form of a hierarchy, deep learning algorithms are effective in producing higher outcomes than using preset and manually produced features. In order to predict the grades of LGG and HGG brain tumors automatically on both whole-brain and just tumor area MRI images, Pereira et al. \cite{pereira2018automatic} proposed an unique Convolutional Neural Network (CNN) with deeper architectures and shorter kernels that achieved 89.5\% accuracy.
Convolutional Neural Networks (CNN) were also utilized for brain tumor classification in \cite{seetha2018brain, bhanothu2020detection, badvza2020classification, das2019brain}.
Seetha and Raja \cite{seetha2018brain} developed a CNN architecture that achieved highest 97.2\% accuracy. Bhanothu et al. \cite{bhanothu2020detection} utilized VGG19 pre-trained model for convolution feature map extraction and used Faster R-CNN as tumor classifier. They achieved mean average precision of 77.6\%. Badza and Barjaktarovic \cite{badvza2020classification} proposed a CNN architecture that achieved highest 96.56\% accuracy, while Das et al. \cite{das2019brain} achieved 94.39\% accuracy with the proposed CNN architecture.
Afshar et al. \cite{afshar2018brain} proposed modified Capsule Network architectures (CapsNets) with five different combinations for brain tumor classification. They used various combination of convolutional layers, convolutional feature maps, varying primary capsules with and without dimensions and varying number of neurons in fully connected layers. The original capsule network achieved 82.30\% accuracy, while CapsNet with one convolutional layer with 64 feature maps achieved the highest accuracy of 86.56\% among all the combinations. 
\begin{table}[!htbp]
\centering
\caption{Summary of some existing machine learning, deep learning and pre-trained architecture based brain tumor detection works.}
\label{tab:reference-sum}
\resizebox{\columnwidth}{!}
{\begin{tabular}{|c|c|c|c|} 
\hline
\textbf{Type}                                                                                    & \textbf{~Reference~} & \textbf{Model used}                                                                                                                 & \textbf{Performance}                                                                                                              \\ 
\hline
\multirow{15}{*}{\begin{tabular}[c]{@{}c@{}}\textbf{Machine }\\\textbf{Learning}\end{tabular}}    & \cite{naik2014tumor}                    & Decision Tree                                                                                                                       & 96\% accuracy                                                                                                                     \\ 
\cline{2-4}& \cite{shil2017improved}                    & SVM                                                                                                                                 & 97.1\% accuracy                                                                                                                   \\ 
\cline{2-4}& \cite{mathew2017tumor}                   & \begin{tabular}[c]{@{}c@{}}DWT for feature extraction \\SVM as classifier\end{tabular}                                              & 99.33\% accuracy                                                                                                                  \\ 
\cline{2-4}& \cite{singh2012classification}                   & \begin{tabular}[c]{@{}c@{}}WT for feature extraction \\SVM as classifier\end{tabular}                                               & 86\%  accuracy                                                                                                                    \\ 
\cline{2-4}& \cite{amin2020distinctive}                   & \begin{tabular}[c]{@{}c@{}}~ GLCM and PCA for feature extraction~~\\SVM-RBF as classifier\end{tabular}                              & \begin{tabular}[c]{@{}c@{}}100\% accuracy using GLCM\\73.07\% accuracy using PCA\end{tabular}                                     \\ 
\cline{2-4}& \cite{ramteke2012automatic}                   & \begin{tabular}[c]{@{}c@{}}KNN \\Linear SVM \\SVM-RBF\end{tabular}                                                                  & \begin{tabular}[c]{@{}c@{}}80\% accuracy using KNN\\~ 67\% accuracy using Linear SVM~~\\69\% accuracy using SVM-RBF\end{tabular}  \\ 
\hline
\multirow{12}{*}{\begin{tabular}[c]{@{}c@{}}\textbf{Deep }\\\textbf{Learning }\end{tabular}}      & \cite{pereira2018automatic}                   & CNN                                                                                                                                 & 89.5\% accuracy                                                                                                                   \\ 
\cline{2-4}& \cite{seetha2018brain}                   & CNN                                                                                                                                 & 97.5\% accuracy \\ 
\cline{2-4}& \cite{bhanothu2020detection}                   & \begin{tabular}[c]{@{}c@{}}VGG16 for feature maps\\Faster R-CNN\end{tabular}                                                        & 77.6\% mean average precision                                                                                                     \\ 
\cline{2-4}& \cite{badvza2020classification}                   & CNN                                                                                                                                 & 96.56\% accuracy                                                                                                                  \\ 
\cline{2-4}& \cite{das2019brain}                   & CNN                                                                                                                                 & 94.39\% accuracy                                                                                                                  \\ 
\cline{2-4}& \cite{afshar2018brain}                   & CapsNets                                                                                                                            & 86.56\% accuracy                                                                                                                  \\ 
\hline
\multirow{22}{*}{\begin{tabular}[c]{@{}c@{}}\textbf{~Pre-trained~}\\\textbf{Models}\end{tabular}} & \cite{khan2020brain}                    & \begin{tabular}[c]{@{}c@{}}VGG16\\ResNet50\\InceptionV3\end{tabular}                                                               & \begin{tabular}[c]{@{}c@{}}96\% accuracy by VGG16\\89\% accuracy by ResNet50\\75\% accuracy by InceptionV3\end{tabular}          \\ 
\cline{2-4}& \cite{swati2019content}                   & VGG19                                                                                                                               & 96.13\% mean average precision                                                                                                    \\ 
\cline{2-4}& \cite{deepak2019brain}                   & GoogLeNet                                                                                                                           & 98\% accuracy                                                                                                                     \\ 
\cline{2-4}& \cite{chelghoum2020transfer}                   & \begin{tabular}[c]{@{}c@{}}AlexNet\\GoogLeNet\\VGG16\\VGG19\\ResNet18\\ResNet50\\ResNet101\\ResNet-Inception-v2\\SENet\end{tabular} & Highest 98.71\% by VGG16                                                                                                          \\ 
\cline{2-4}& \cite{mehrotra2020transfer}                   & \begin{tabular}[c]{@{}c@{}}AlexNet\\GoogLeNet\\ResNet50\\ResNet101\\SqueezeNet\end{tabular}                                         & Highest 99.04\% by AlexNet                                                                                                        \\
\hline
\end{tabular}}
\end{table}

\noindent\textbf{(c) \underline{Pre-trained Models}}:
Although recent deep learning models have demonstrated good classification performance, they still require large datasets. Several studies have demonstrated that employing pre-trained models improves effectiveness in detecting brain tumors. To classify brain tumors from MRI images, Khan et al. \cite{khan2020brain} employed VGG16, ResNet50, and InceptionV3 models, whereas Swati et al. \cite{swati2019content} used a VGG19 based feature extractor. Deepak and Ameer \cite{deepak2019brain} introduced a classification method that employed deep transfer learning and a pre-trained GoogLeNet architecture to extract features from brain MRI images with five fold cross validation to categorize three types of brain tumors: glioma, meningioma, and pituitary. On the same dataset, Chelghoum et al. \cite{chelghoum2020transfer} employed AlexNet, GoogLeNet, VGG16, VGG19, ResNet18, ResNet50, ResNet101, ResNet-InceptionV2 and SENet pre-trained models which they trained for varying number of epochs to investigate the correlation between time and model accuracy. Mehrota et al. \cite{mehrotra2020transfer} performed a binary classification task of detecting malignant and benign tumors. With just $696$ T1-weighted MRI images in the dataset, they employed pre-trained models such as AlexNet, GoogLeNet, ResNet50, ResNet101 and SqueezeNet.

\subsection{Explainable Artificial Intelligence}
Many Explainable Artificial Intelligence approaches have already been presented for image classification and understanding tasks, but little emphasis has been dedicated to explain brain imaging such as brain tumor identification, segmentation tasks. For the objective of making the proposed models more interpretable, 2D Grad-CAM was added to \cite{natekar2020demystifying, esmaeili2021explainable,  windisch2020implementation}. Natekar et al. \cite{natekar2020demystifying} used 2D Grad-CAM to explain the predictions of the Deep Neural Network (DNN) on brain tumor classification. While Esmaeili et al. \cite{esmaeili2021explainable} used 2D Grad-CAM for performance comparison among DenseNet-121, GoogLeNet and MobileNet on brain tumor classification, Windisch et al. \cite{windisch2020implementation} generated heatmaps from 2D GRAD-CAM to indicate the areas of the existence of a brain tumor predicted by the proposed model. Class activation mapping (CAM) is extended in Saleem et al. \cite{saleem2021visual} by creating 3D heatmaps to illustrate the significance of segmentation data. Zeineldin et al. \cite{zeineldin2022explainability} introduced a new framework (NeuroXAI) for visualizing deep learning networks by implementing seven cutting-edge ways of explanation. They used the magnetic resonance (MR) modality to classify and segment brain tumors using NeuroXAI.

\subsection{Cost-Sensitive Deep Neural Networks}
In order to train cost-sensitive neural networks, Zhou and Liu \cite{zhou2005training} examined the effects of over-sampling, under-sampling, threshold-moving, hard-ensemble, soft-ensemble, and SMOTE on $21$ data sets with three different types of cost metrics and a real-world cost-sensitive data set. According to their findings, cost-sensitive learning is reasonably simple for binary classification tasks but challenging for multi-class classification tasks. A sequential three-way decision model based on deep neural network was suggested by Li et al. \cite{li2017cost} utilizing a non-linear feature extraction technique. In the training phase of the deep neural network, sequential granular features are produced based on many iterations in order to balance the misclassification and test costs. A cost-sensitive sequential three-way decision model technique is used based on variations in the hierarchical granular structure of the input images. Their tests confirmed the efficacy of the sequential 3WD technique and the use of granular features in the input images. A comprehensive approach to the issue of small datasets was proposed by Xu et al. \cite{xu2020mscs} by employing an ensemble of several lightweight networks. According to the experimental findings, shallower networks outperformed deeper ones, however transfer learning performed poor in case of 3D Convolutional Neural Networks. In order to address the issue of class imbalance, they also proposed a new loss function that combines cross-entropy with an Area under the ROC Curve (AUC) approximation.

\section{Dataset Description}\label{sec3}
\begin{figure}[t]
		\centering
		\begin{subfigure}[b]{.20\columnwidth}
		    \centering
			\includegraphics[width=1\linewidth]{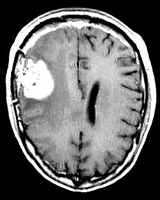}
                \subcaption{}
			\label{fig:sample_BR35H_a}
		\end{subfigure}
		\begin{subfigure}[b]{.20\columnwidth}
		    \centering
			\includegraphics[width=1\linewidth]{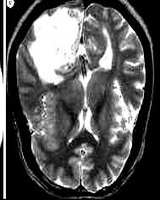}
                \subcaption{}
                \label{fig:sample_BR35H_b}
		\end{subfigure}
        \begin{subfigure}[b]{.20\columnwidth}
		    \centering
			\includegraphics[width=1\linewidth]{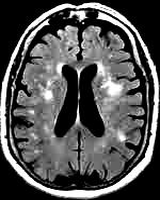}
                \subcaption{}
                \label{fig:sample_BR35H_c}
		\end{subfigure}
		\begin{subfigure}[b]{.20\columnwidth}
		    \centering
			\includegraphics[width=1\linewidth]{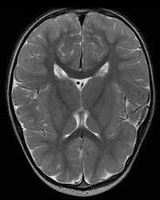}
                \subcaption{}
                \label{fig:sample_BR35H_d}
		\end{subfigure}
		
		\caption{Few sample instances of \textbf{Tumor (a \& b)} and \textbf{No-Tumor (c \& d)} category from \textit{Br35H} dataset.}
		\label{fig:sample_BR35H}
\end{figure}

We utilized two different datasets for brain tumor detection tasks. One of them is \textit{Br35H :: Brain Tumor Detection 2020}\footnote{\url{https://www.kaggle.com/datasets/ahmedhamada0/brain-tumor-detection}} which has two category. One is the yes (Tumor) category and another is the no (No-Tumor) category. The class wise data distribution of the datasets are presented in Table \ref{tab:dataset-frequency}. The dataset is balanced as both the classes of Tumor and No-Tumor have $1500$ instances. Some samples of the dataset from both the category are illustrated in Figure \ref{fig:sample_BR35H} that shows the samples from both the category where \ref{fig:sample_BR35H_a} and \ref{fig:sample_BR35H_b} show the presence of the tumor and Figure \ref{fig:sample_BR35H_c} and \ref{fig:sample_BR35H_d} show the samples from No-Tumor category which do not have any tumor in the brain MRI images. 
\begin{table}[h]
		\centering
            \caption{Class-wise data distribution on both datasets.}
		\label{tab:dataset-frequency}
		\begin{tabular}{ccc}
			\toprule
			\textbf{Dataset} & \textbf{Tumor}  & \textbf{No-Tumor} \\
			\midrule
			Br35H & $1500$ & $1500$\\
			Brain MRI Images for Brain Tumor Detection (BTD) & $155$ & $98$\\
			\bottomrule
		\end{tabular}
\end{table}

\begin{figure}[t]
		\centering
		\begin{subfigure}[b]{.20\columnwidth}
		    \centering
			\includegraphics[width=1\linewidth]{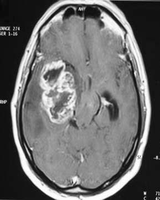}
                \subcaption{}
                \label{fig:sample_detection_a}
		\end{subfigure}
		\begin{subfigure}[b]{.20\columnwidth}
		    \centering
			\includegraphics[width=1\linewidth]{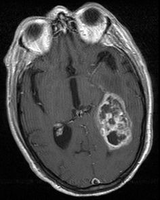}
                \subcaption{}
                \label{fig:sample_detection_b}
		\end{subfigure}
        \begin{subfigure}[b]{.20\columnwidth}
		    \centering
			\includegraphics[width=1\linewidth]{./36n.png}
			\subcaption{}
                \label{fig:sample_detection_c}
		\end{subfigure}
		\begin{subfigure}[b]{.20\columnwidth}
		    \centering
			\includegraphics[width=1\linewidth]{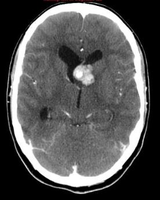}
                \subcaption{}
                \label{fig:sample_detection_d}
		\end{subfigure}
		\caption{Few sample instances of \textbf{Tumor (a \& b)} and \textbf{No-Tumor (c \& d)} category from \textit{Brain MRI Images for Brain Tumor Detection} dataset.}
		\label{fig:sample_detection}
	\end{figure}

\noindent The second dataset used in this study is \textit{Brain MRI Images for Brain Tumor Detection}\footnote{\url{https://www.kaggle.com/datasets/navoneel/brain-mri-images-for-brain-tumor-detection}}. It has also two classes which are yes (Tumor) and no (No-Tumor). Tumor class has $155$ MRI images which contains brain tumor and (No-Tumor) class has $98$ MRI images of the healthy brain. The size of the dataset is not so large. Some sample images from this dataset are presented in Figure \ref{fig:sample_detection} which contains the samples from both the category where \ref{fig:sample_detection_a} and \ref{fig:sample_detection_b} have tumor in the brain MRI images and Figure \ref{fig:sample_detection_c} and \ref{fig:sample_detection_d} show the samples from No-Tumor category. 

\section{Background Study} \label{sec4}
The inner workings of a neural network is a black box to us. However, we can visualize the hidden layers of a neural network using a variety of Explainable AI techniques. Some XAI strategies for explaining the output of a trained model include LIME, SHAP and CAM. Some options for explaining gradient-based neural network models are Grad-CAM, Grad-CAM++, Score-CAM, Faster Score-CAM, Vanilla Saliency and SmoothGrad. LIME is another explainable technique that we can use for explaining a model where gradient-based weights are generated. We may use these techniques to explain the outputs of the standard CNN and other pre-trained CNN architectures to detect brain tumors. This section depicts the background studies of all the neural network models used in our experiments and the explainable AI techniques used to interpret the model's behavior.  


\subsection{Cost-Sensitive Neural Network}
Data is the most salient factor of a neural network. From basic to advance, all the neural network architectures are data driven. The performance of these models are heavily relied on data. In case of classification tasks, if the number of data from all the classes are not equal then these neural network based models exhibit poor performance. Because during training, more weight updation is performed for the class which has more data and less updation of weight occurs for the class which has limited data. To fix this biasness, we can implement cost sensitive learning. With the help of cost-sensitive neural network, also known as weighted neural network, the learning algorithm like back-propagation can be modified to account for misclassification errors according to the significance of the class. Utilizing this technique in case of an imbalanced dataset, we can force the model to pay more attention to samples from the minority class than the majority class by implying hand-crafted weight for each class to the cost function that is being minimized. Equation \ref{eqn:logloss} represents default log loss function that can be optimized for classification task. If we use this equation in a classification task of an imbalanced data, it will put equal weight to the two classes which can cause potential bias.

\begin{equation}  
LogLoss=\frac{1}{N} \sum_{i=1}^N\left[-\left(y_i* \log \left(\widehat{y_i}\right)+\left(1-y_i\right)* \log \left(1-\widehat{y_i}\right)\right)\right.
\label{eqn:logloss}
\end{equation}

\begin{equation}
WeightedLogLoss =\frac{1}{N} \sum_{i=1}^N\left[-\left(w_0\left(y_i* \log \left(\widehat{y_i}\right)\right)+w_1\left(\left(1-y_i\right)* \log \left(1-\widehat{y_i}\right)\right)\right)\right]
\label{eqn:weightedlogloss}
\end{equation}

\noindent Instead for the classification task on an imbalanced dataset, equation \ref{eqn:weightedlogloss} can be used where we can put appropriate class weights according to the class label distribution. If small class weight is applied to the cost function for the majority class, it will generate a small error which will cause less update to the trainable weights of the model. And in contrast, if large class weight is used for minority class, the model will produce more error and thus more updation of trainable weights of the model will be required.

\subsection{Gradient-based Explainable AI}
The partial derivative of the output predictions across each layer of the neural network with regard to the input images is calculated using gradient-based XAI algorithms, which are commonly used to produce feature attribution maps. Some of the gradient-based XAI methods used in this study are discussed below.

\noindent\textbf{(a) \underline{Grad-CAM}}: 
Grad-CAM, or gradient-weighted class activation maps, is an acronym. There are several ways that Grad-CAM is different from conventional CAM. Traditional CAM, which passes the convolutional feature space directly to the output layer, can be used with small ConvNets that don't have any thick layers \cite{selvaraju2017grad}. Grad-CAM, on the other hand, is founded on saliency maps, which show which pixels in an input image are relevant. Prior to replacing it in the implementation with a linear function, Grad-CAM computes the gradient of the output layer the class prediction layer—with respect to the feature maps of the previous convolutional layer. Gradients then flow back to determine the relative significance of these feature maps so that average pooling can be used to predict classes. A gradient-weighted CAM heatmap depicting positive and negative key variables for the source images can be produced after establishing a weighted combination of the feature maps and their weights. The region of attention is most likely to be found in those areas. The heatmap was then subjected to a ReLU function to eliminate the negative areas, reduce their significance to zero, and keep just the relevance of the positive areas. Equation \ref{eqn:grad1} and \ref{eqn:grad2} define the idea more formally. 
\begin{equation}
G r a d C a m_i^c=R e L U\left(\sum_i \alpha_i^c A^i\right)
\label{eqn:grad1}
\end{equation}

\begin{equation}
\alpha_i^c=\frac{1}{N} \sum_x \sum_y \frac{\partial y^c}{\partial A_{x, y}^i}
\label{eqn:grad2}
\end{equation}
$\alpha_i^c$ denotes the gradient of any target class c, an individual layer $i$'s activation feature map, $M^i$, is calculated by averaging the magnitude of the spatial location and N is number of pixels. 

\noindent\textbf{(b) \underline{Grad-CAM++}}:
Grad-CAM++, a class activation map where the flaws of Grad-CAM have been dealt with a better understanding of CNN based models. Grad-CAM is not particularly accurate when it comes to encircling the class region in an image. Chattopadhay et al. \cite{chattopadhay2018grad} proposed a better version of Grad-CAM where heatmaps are generated throughout all the regions of a class if the prediction occurs on the basis of multiple locations of an image. As a result, when several instances of a particular class are present in the images, the produced heatmap depicts the model's behavior more clearly. They improvised the neuron importance weights ($\alpha_i^c$) of Grad-CAM from \ref{eqn:grad2} and proposed the new one where global pooling layers are missing which can be seen in Equation \ref{eqn:Grad-Cam++}. 
\begin{equation}
\alpha_i^c=\sum_x \sum_y w_{x y}^{i c} \cdot R e L U\left(\frac{\partial y^c}{\partial A_{x,y}^i}\right)
\label{eqn:Grad-Cam++}
\end{equation}
Here, $w_{x y}^{i c}$ is the gradient weights for a specific class $c$ and activation map $i$ for a particular spatial location $x, y$.

\noindent\textbf{(c) \underline{Saliency Map}}:
Deep image recognition ConvNets are visualized using saliency maps, which were trained on the enormous ImageNet challenge dataset. Simonyan et al. \cite{simonyan2013deep} showed that by numerically optimizing the input image, it is possible to create clear representations of ConvNet classification models. The visualization approach entails mathematically producing an image that, according to the ConvNet class score model, represents the class given a learned classification ConvNet and a class of relevance. 
\begin{equation}
\arg \max _I S_c(I)-\lambda\|I\|_2^2
\label{eqn:saliency}
\end{equation}
Saliency map tries to find a regularised image on the basis of Ridge Regression (L2) by which the score \textbf{Sc(I)} of class \textbf{c} is always high. Equation \ref{eqn:saliency} describes the process for visualizing the class models where $\lambda$ is the parameter for regularization. 

\noindent\textbf{(d) \underline{SmoothGrad}}:
In addition to being compatible with different sensitivity map techniques, SmoothGrad \cite{smilkov2017smoothgrad} tends to minimize visual noise. The average is determined after adding Gaussian noise and computing the gradient for several samples encircling the given Data. A mathematical formulation of the saliency map can be found in Equation \ref{eqn:SmoothGrad}.

\begin{equation}
\overline{M_c}\left(X^i\right)=\frac{1}{N} \sum_1^N M_c\left(X^i+G\left(0, \sigma^2\right)\right)
\label{eqn:SmoothGrad}
\end{equation}
Here, $N$ is the frequency of data, $M_c(X^i)$ is the mean of the final sensitivity maps, and $G(0, \sigma^2)$ denotes Gaussian noise having variance $\sigma^2$. 

\noindent\textbf{(e) \underline{Score-CAM}}:
Wang et al. \cite{Wang_2020_CVPR_Workshops} proposed a finer explanation technique for visualizing the significant region of a classification model, which they call Score-weighted Class Activation Mapping or Score-CAM. They addressed the shortcomings of gradient-based visualization methods like CAM \cite{Zhou_2016_CVPR}, Grad-CAM \cite{selvaraju2017grad} and the further improvement of Grad-CAM++ \cite{chattopadhay2018grad} and suggested a method for deriving the significance of activation map from the subsidy of the underlined input features to the output of the model where other visualization models discussed earlier use the spatial sensitivity (gradient information). Equation \ref{eqn:Score-CAM} clarify the idea of Score-CAM if we compare it with equation \ref{eqn:grad2} and \ref{eqn:Grad-Cam++}.
\begin{equation}
\alpha_i^c=C\left(A^i\right)
\label{eqn:Score-CAM}
\end{equation}
Here, $C(*)$ is the $Channel-wise Increase of Confidence$ which is used for measuring the significance of each of the activation maps.

\noindent\textbf{(f) \underline{Faster Score-CAM}}:
A faster variation of Score-weighted Class Activation Mapping (Score-CAM) \cite{Wang_2020_CVPR_Workshops}, sometimes known as "Faster Score-CAM". It is more efficient than Score-CAM and explains the model more precisely. In Faster Score-CAM, the channels with substantial variations are considered as mask images, given that different image channels play a significant role in creating the final heat map\footnote{\url{https://github.com/tabayashi0117/Score-CAM/blob/master/README.md\#fasterscore-cam}}.

\subsection{Perturbation-based Explainable AI: LIME}
Through forward training of the model, perturbation-based approaches examine the network by altering the input features and calculating the effect on the output predictions. Ribeiro et al. \cite{ribeiro2016should} proposed Local Interpretable Model-Agnostic Explanations (LIME) which describes the predictions of any classifier in an explicable manner, by acknowledging an interpretable model locally around the prediction. The primary function provided by LIME splits the input image into regions, which are then allocated saliency weights that configures the foundation to produce explanation. These saliency weights are calculated based on the level to which the presence or absence of that region of the image affects the classification result of the underlying model. XAI pattern produced by LIME depicts that the green marked regions of the images indicate that the neural networks used this region to correctly classify images while the red marked regions are the wrongly labeled portion by the neural network models. Equation \ref{eqn:lime} defines the idea more formally.

\begin{equation}
interpretation(x) = arg\;min_{v\epsilon V} L(u, v, \pi_x) + \omega(v)
\label{eqn:lime}
\end{equation}
\\
Here $v$ is the explainable model for the sample $x$ which will reduce loss $L$ and measures how close the explanation is, in comparison to the predicted value of the initial model $u$. Here, $V$ is the group of realizable explanations. The closeness measure $\pi_x$ interprets the range of the locality around sample $x$, and that is what we consider for the explanation. 

\subsection{Performance Metrics}
Accuracy, Precision, Recall, F1-Score and Specificity are the performance metrics employed in this work. The ratio of accurately predicted samples over all samples is calculated to determine accuracy. The metric where false positive cases are given more significance is precision or positive predictive value. The metric called Recall or Sensitivity is used to compute the true positive rate and gives false negative cases more weight. The F1-score is the harmonic mean of recall and precision that equally weights false positive and false negative cases. The likelihood of a negative test result, presuming the subject is actually negative, is known as specificity (true negative rate).

\section{Proposed Methodology}\label{sec5}
\begin{figure}[h]
		\centering
			\includegraphics[width=1\linewidth]{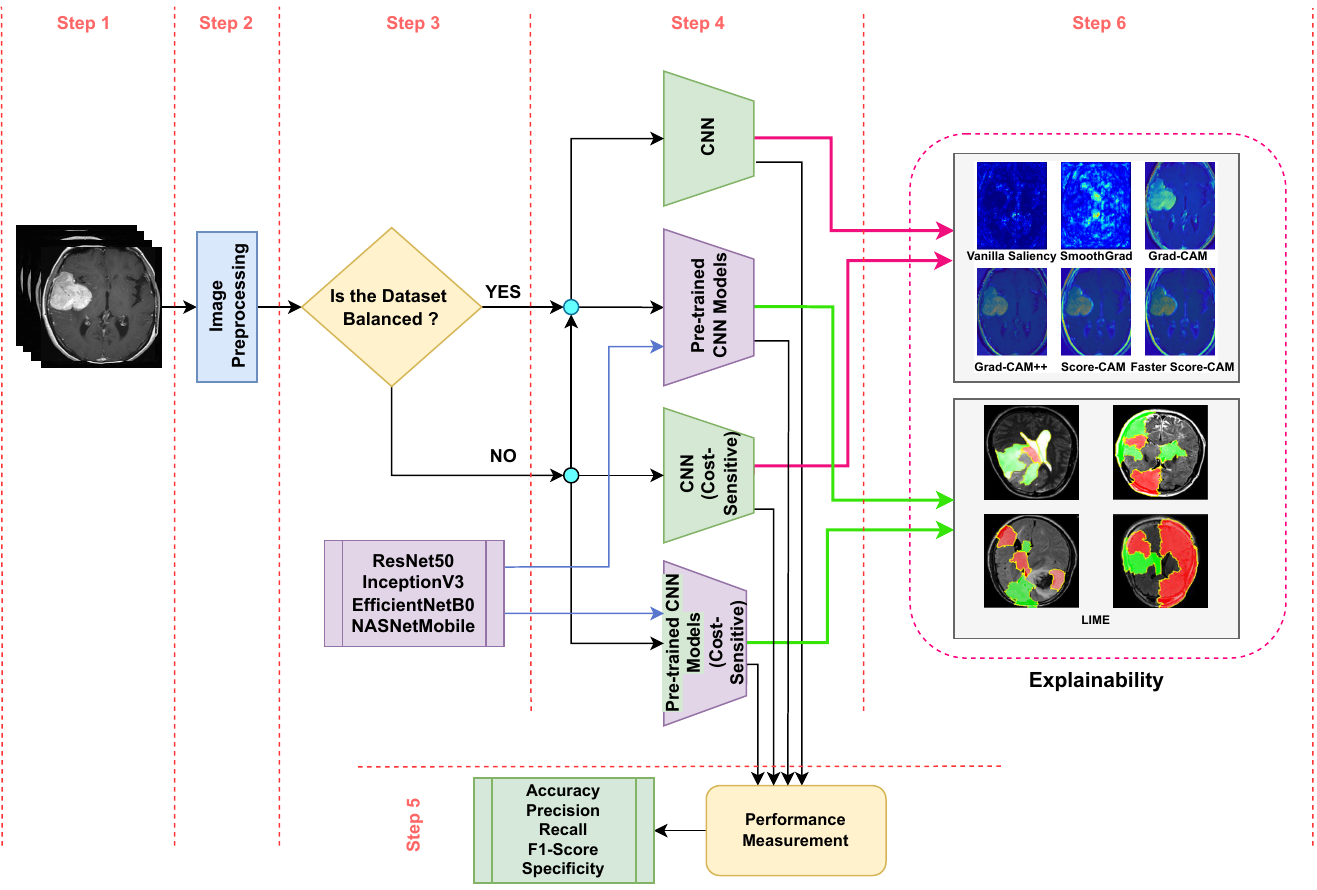}
		\caption{Schematic diagram of the proposed brain tumor detection system.}
		\label{fig:methodology}
	\end{figure}

This section represents the workflow of our proposed approach where some sample images are presented as input and the explainations are shown for some random samples from the dataset for the illustration purpose. Figure \ref{fig:methodology} depicts the methodology of our study.  
\begin{enumerate}[label=\textbf{Step \arabic{enumi})}, wide, labelwidth=!, labelindent=0pt]
    \item \textbf{Image Input}: MRI images of Brain were first loaded for the further pre-processing before presenting to the model. 
    \item \textbf{Image Preprocessing}: The loaded images were passed through a resizing step where a bi-linear interpolation method was used as the size of the images were not same. We resized the images of the dataset to $224 * 224$. Then, the datasets were split into train, validation and test set having the proportion of 80\%, 10\% and 10\% respectively. We kept 80\% of the data for training considering the small number of data in both the datasets. 
    \item \textbf{Check Data Equivalence}: We now compare the data equivalence of the two classes to determine the process for the next step.
    
    \item \textbf{Train Classification Models}: 
    \begin{itemize}
        \item Standard CNN and four pre-trained CNN models (ResNet50, InceptionV3, EfficientNetB0, and NASNetMobile) are trained if the data is balanced.
        \item If the dataset is imbalanced we use a cost sensitive neural network technique where the model pays more importance to minority classes by introducing hand-crafted weight. For a better comparison, a simple convolutional neural network and four pretrained CNN models are also used here with an additional weight to the minority class.
    \end{itemize}
    
    \item \textbf{Performance Evaluation}: The evaluation of the model is performed here with five different performance metrics which are accuracy, precision, recall, f1-score, and specificity.
    \item \textbf{Explainability}: The prediction of the models are explained here with various explainable AI methods. 
    \begin{itemize}
        \item We considered Vanilla Saliency, SmoothGrad, Grad-CAM, Grad-CAM++, Score-CAM and Faster Score-CAM for explaining the CNN based model.
        \item LIME is used for the explanation of the pre-trained models. The details of these methods can be found in section \ref{sec4}.
    \end{itemize}
    
\end{enumerate}
	
\section{Model Architecture} \label{sec6}
\begin{figure*}[h]
		\centering
			\includegraphics[width=1\linewidth]{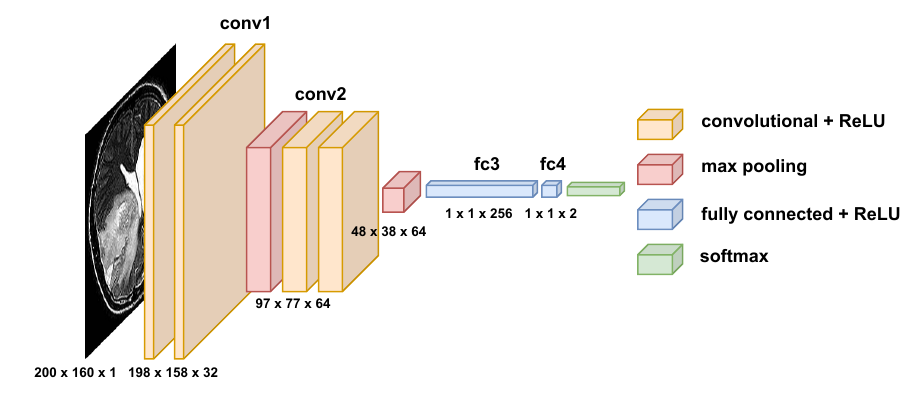}
		\caption{Architecture of the standard Convolutional Neural Network (CNN) model.}
		\label{fig:cnnarch}
	\end{figure*}

This section describes the detailed architecture of the models utilized in the proposed methodology.\\
\noindent\textbf{(a) \underline{Convolutional Neural Network}}:
It is a neural network that, in essence, functions as a feature extractor and has been demonstrated to be especially effective at image recognition, processing, and classification \cite{albawi2017understanding}. CNN and artificial neural networks work pretty similarly. A simple structure in an image can be detected through convolution. In order to determine training weights and evaluate its performance, CNN needs training data and validation. Each input image is processed using a series of convolution layers containing filters (Kernels): Convolution layers, Pooling layers, and Fully-connected layer (FC), all of which employ the sigmoid function. The sigmoid function is used to classify an item using a probabilistic value, which in binary classification is either 0 or 1. A fully connected layer functions as a classifier, while the Convolution $+$ MaxPooling layers retrieve features from the input image. In the example above, when the network receives an image as input, it assigns the highest probability to it and predicts which class the image belongs to. Our CNN based model contains $2$ convolution layer which are the size of $32$ and $64$ neuron and a dense layer of $256$ neurons. Here a $3 \times 3$ kernel is used. The activation function used here is the Rectified Linear Unit (ReLU). In neural networks, especially CNNs, ReLU is the most widely employed activation function. It converges more quickly. When $x$ grows large enough, the slope does not plateau or saturate. Other activation functions, such as sigmoid and tanh, suffer from the problem of vanishing gradients. It's only lightly activated. Because ReLU is zero for all negative inputs, any given unit is likely to remain inactive. This is frequently desirable. We also used $25\%$ dropout in our model to avoid excessive and unnecessary computation. The architecture of our CNN based model can be seen in Figure \ref{fig:cnnarch}.

\noindent\textbf{(b) \underline{ResNet50}}:
ResNet50 is a very widely used deep residual network with 50 layers consisting of 3-layer bottleneck blocks \cite{he2016deep} having 3.8 billion floating point operations per second. It uses the idea of skip-connection permitting the model to move forward without experiencing a significant change ignoring the vanishing gradient problem completely. In our work, we fine-tuned all the layers of the pre-trained models followed by 3 consecutive dense layers having the size of $1024$, $512$ and $2$ respectively. ReLU activation function is used for the hidden layers, and Softmax is used at the top layer. There is also a drop layer having the rate of $0.5$ in the model to overcome the over-fitting problem. 

\noindent\textbf{(c) \underline{InceptionV3}}:
InceptionV3 is a convolutional neural network \cite{szegedy2016rethinking} that was designed to improve the performance of image classification on the ImageNet Large Scale Visual Recognition Challenge (ILSVRC). It starts with a stem module that extracts low-level features from the input image. The core of the network is a series of Inception modules. Each module is composed of multiple parallel convolutional layers with different filter sizes, including 1x1, 3x3 and 5x5 convolutions. This allows the network to capture features at different scales and preserve spatial information. InceptionV3 also includes reduction modules that reduce the spatial dimensions of feature maps. This helps reduce computational complexity and introduces more abstract and high-level features. The network also incorporates auxiliary classifiers at intermediate layers during training. These classifiers provide additional supervision signals and help combat the vanishing gradient problem. During inference, these auxiliary classifiers are discarded. At the end of the network, global average pooling is applied to generate a fixed-size feature vector. This vector is then fed into fully connected layers for classification or other downstream tasks.

\noindent\textbf{(d) \underline{EfficientNetB0}}:
EfficientNetB0 is a pre-trained deep convolutional neural network that is part of the EfficientNet \cite{tan2019efficientnet} model family developed through neural architecture search techniques. The architecture of EfficientNetB0 follows a compound scaling approach, which uniformly scales the network's depth, width, and resolution. It consists of stacked blocks, each containing a combination of depth-wise separable convolutions, bottleneck structures, and squeeze-and-excitation modules. The fundamental building block of EfficientNetB0 is the depth-wise separable convolution, which separates spatial and channel-wise operations. This reduces computational complexity while preserving the network's representational power. The bottleneck structure further improves efficiency by reducing the number of parameters. Moreover, EfficientNetB0 incorporates squeeze-and-excitation modules, which dynamically re-calibrate channel-wise feature responses using learned attention weights. This enhances the network's focus on informative features and boosts its discriminative capabilities. The network architecture is characterized by stacked blocks, with each block having a specific number of repetitions. The scaling coefficients, which control the network's size, determine the number of repetitions and overall depth. These coefficients are selected to optimize the trade-off between model size and accuracy.

\noindent\textbf{(e) \underline{NASNetMobile}}:
NASNetMobile is a pre-trained neural network known as Neural Architecture Search Network Mobile \cite{zoph2018learning} and constructed using neural architecture search techniques. The design of the NASNetMobile model is based on a cell-based architecture, where the cell structure is replicated multiple times to create the entire network. Each cell contains a series of operations that are applied to the input feature maps. The architecture comprises two types of cells: normal cells and reduction cells. Normal cells in the NASNetMobile model are responsible for capturing spatial features at different resolutions while preserving the spatial dimensions of the feature maps. These cells incorporate various operations such as convolutions, pooling, and skip connections. The selection of these operations is determined through a neural architecture search process that utilizes reinforcement learning techniques to identify the most effective ones. On the other hand, reduction cells are designed to decrease the spatial dimensions of the feature maps while increasing the number of channels. They contribute to reducing the computational complexity of the network. Reduction cells typically employ larger strides and incorporate pooling operations to down sample the feature maps. The overall architecture of NASNetMobile is created by stacking the normal and reduction cells in an alternating pattern. The specific number and arrangement of cells are determined during the neural architecture search process to strike a balance between model size and accuracy. The final architecture is chosen based on its performance on a given task or dataset. NASNetMobile is designed to be lightweight and efficient, making it well-suited for deployment on mobile and embedded devices that have limited computational resources.
\begin{table}[]
\centering
\caption{Hyper-parameter settings to train/fine-tune the models in the proposed methodology. Here, CS denotes Cost Sensitive.}
\label{tab:hyper}
\begin{tabular}{ccccc}
\toprule
\textbf{Dataset}                                                                                                    & \textbf{Model}                   & \textbf{Epoch} & \textbf{Batch Size} & \textbf{Learning Rate}\\ \midrule
\multirow{2}{*}{Br35H}                                                      & CNN                     &   7   &    50  & $1e^{-3}$   \\ 
& ResNet50                &    20    &  8   &  $1e^{-5}$  \\
        & InceptionV3                &   20     &  8   &  $1e^{-5}$ \\
        & EfficientNetB0                &   29     &   8  &  $1e^{-5}$  \\
        & NASNetMobile         &   33     &  8   & $1e^{-5}$  \\
       \midrule
\multirow{10}{*}{\begin{tabular}[c]{@{}c@{}}Brain MRI \\ Images for\\ Brain Tumor\\ Detection\end{tabular}} & CNN                     & 15       & 50     & $1e^{-3}$    \\ 
        & ResNet50                &   35     &  .8   & $1e^{-5}$   \\
        & InceptionV3                &   30     &  8   &  $1e^{-5}$ \\
        & EfficientNetB0                &   35     &   8  &  $1e^{-5}$ \\
        & NASNetMobile                &    48    &   8  & $1e^{-4}$  
          \\ \cline{2-5} 
        & CS-CNN      &     35   &  50   & $2e^{-5}$   \\ 
        & CS-ResNet50                &    80    &  8   & $1e^{-5}$  \\
        & CS-InceptionV3                &      120     &  8   & $1e^{-4}$ \\
        & CS-EfficientNetB0                &    80    &  8   & $2e^{-7}$  \\
        & CS-NASNetMobile                &   110     &  8   & $2e^{-7}$ \\ \midrule
\end{tabular}
\end{table}


\section{Experiments}\label{sec7}

\noindent \textbf{(a) \underline{Experimental Setup}}:
We have used Python $3.6.5$ and the versions used for different libraries are as follows: Tensorflow $2.10.0$, keras $2.10.0$ and LIME $0.2.0.1$. Keras-vis library is used for performing other XAI techniques (Vanilla Saliency, SmoothGrad, Grad-CAM, Grad-CAM++, Score-CAM and Faster Score-CAM). Most of the experiments have been conducted on jupyter notebook, and some are performed in Google Colaboratory\footnote{\url{https://colab.research.google.com}}.\\
\noindent\textbf{(b) \underline{Hyper-parameter Settings}}:
Both the \textit{Br35H} and \textit{Brain MRI Images for Brain Tumor Detection (BTD)} datasets are run through a convolutional neural network and 4 pre-trained CNN models (ResNet50, InceptionV3, EfficientNetB0, and NASNetMobile). The imbalanced dataset (BTD) is also used to perform the cost-sensitive models. Cross Entropy loss is used as the loss function which is the average of the sum of the log of improved forcasted probabilities for each data, which can be defined using the equation \ref{eqn:logloss}. The epoch, optimizer, learning rate and batch size for each of the models can be seen in Table \ref{tab:hyper}. Both the datasets were divided into training, validation and test sets, containing $80\%$, $10\%$, $10\%$ of the images respectively. Table \ref{tab:split} shows the train, validation and test split of both the dataset.
\begin{table}[]
\centering
\caption{Train-validation-test split for both datasets.}
\label{tab:split}
\begin{tabular}{ccccc}
\hline
\textbf{Dataset}                                                                          & \textbf{Class} & \textbf{Train} & \textbf{Validation} & \textbf{Test} \\ \hline
\multirow{2}{*}{Br35H}                                                                    & Tumor          & 1200           & 150                 & 150           \\ \cline{2-5}& No-Tumor       & 1200           & 150                 & 150           \\ \hline
\multirow{3}{*}{\begin{tabular}[c]{@{}c@{}}Brain MRI Images for \\Brain Tumor Detection\end{tabular}} & Tumor          & 124            & 15                  & 16            \\ \cline{2-5}& No-Tumor       & 78             & 10                  & 10            \\ \hline
\end{tabular}
\end{table}
			      
\begin{figure}[h]

        \centering
		\begin{subfigure}[b]{.28\columnwidth}
		    \centering
			\includegraphics[width=1\linewidth]{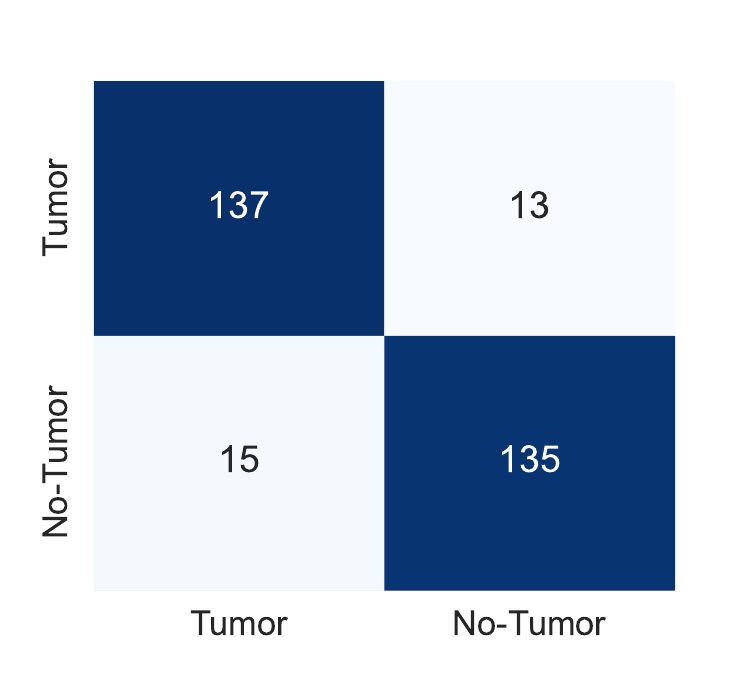}
            \subcaption{CNN}
            \label{ResBr35H}
		\end{subfigure}
		\begin{subfigure}[b]{.28\columnwidth}
		    \centering
			\includegraphics[width=1\linewidth]{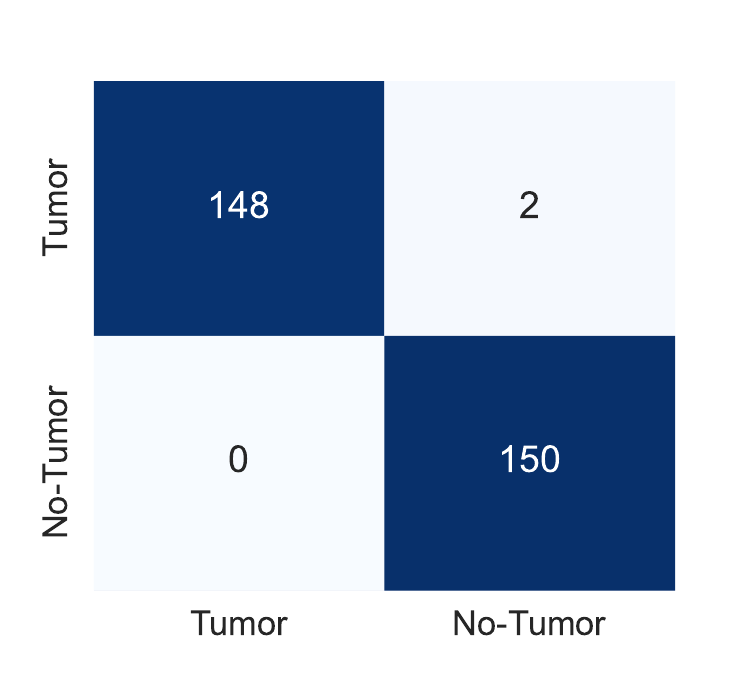}
            \subcaption{ResNet50}
            \label{CNNBr35H}
		\end{subfigure}
		\begin{subfigure}[b]{.28\columnwidth}
		    \centering
			\includegraphics[width=1\linewidth]{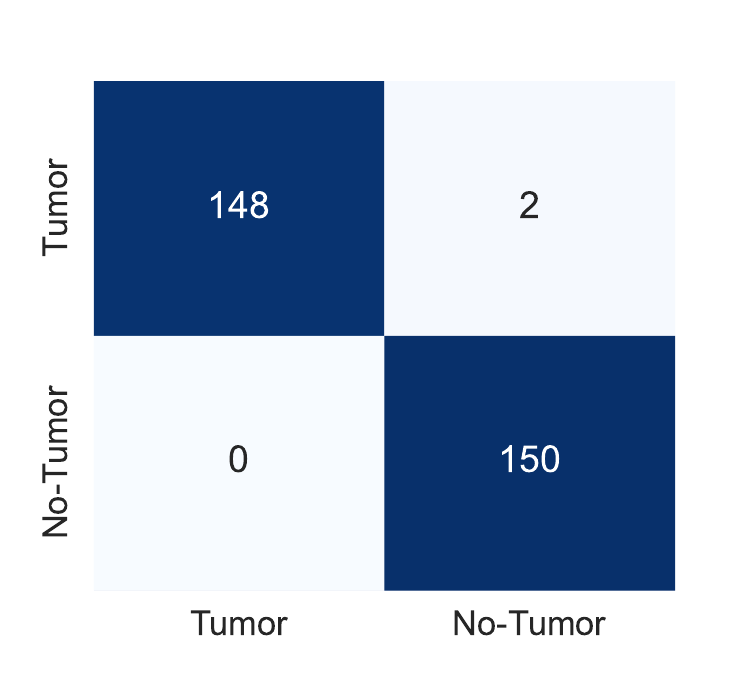}
            \subcaption{InceptionV3}
            \label{ResBTD}
		\end{subfigure}
		\begin{subfigure}[b]{.28\columnwidth}
		    \centering
			\includegraphics[width=1\linewidth]{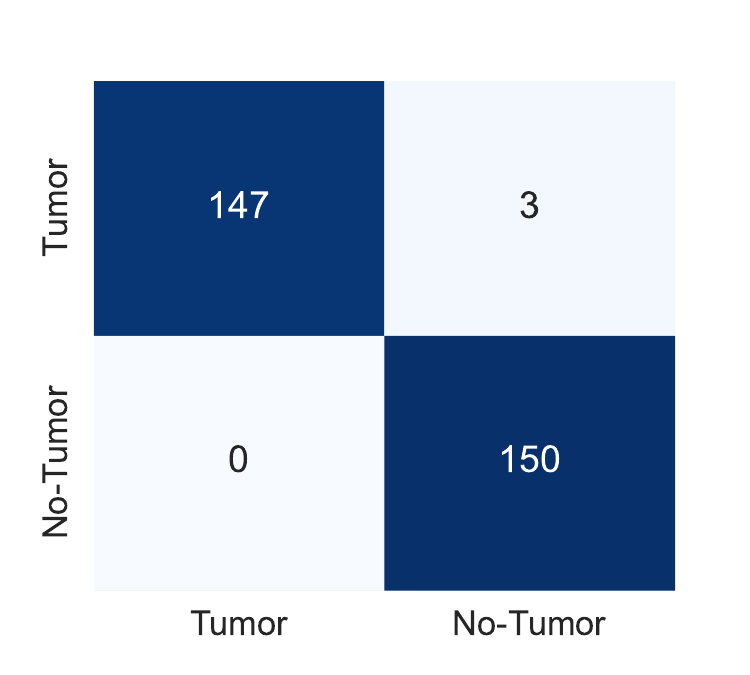}
            \subcaption{EfficientNetB0 }
            \label{CNNBTD}
		\end{subfigure}
            \begin{subfigure}[b]{.28\columnwidth}
		    \centering
			\includegraphics[width=1\linewidth]{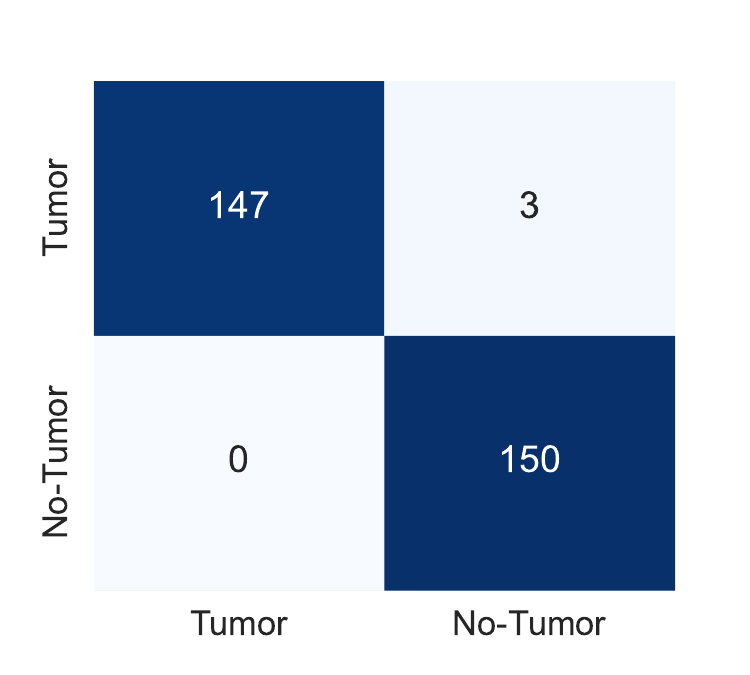}
            \subcaption{NASNetMobile}
            \label{CSResBTD}
		\end{subfigure}
            
		\caption{Confusion matrices of all the models on \textit{Br35H} dataset.}
		\label{fig:confusion_Br}
	\end{figure}

\begin{figure}[h]

        \centering
		\begin{subfigure}[b]{.28\columnwidth}
		    \centering
			\includegraphics[width=1\linewidth]{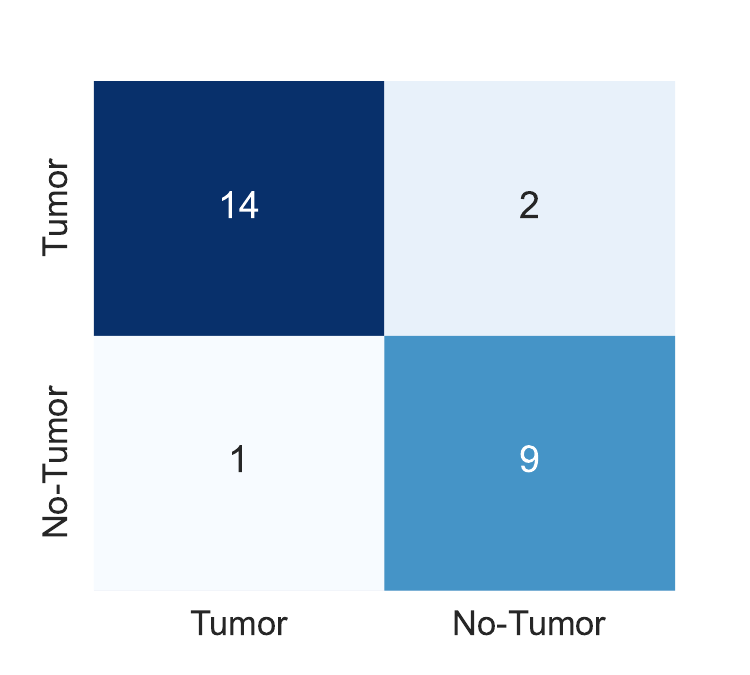}
            \subcaption{CNN}
            \label{ResBr35H}
		\end{subfigure}
		\begin{subfigure}[b]{.28\columnwidth}
		    \centering
			\includegraphics[width=1\linewidth]{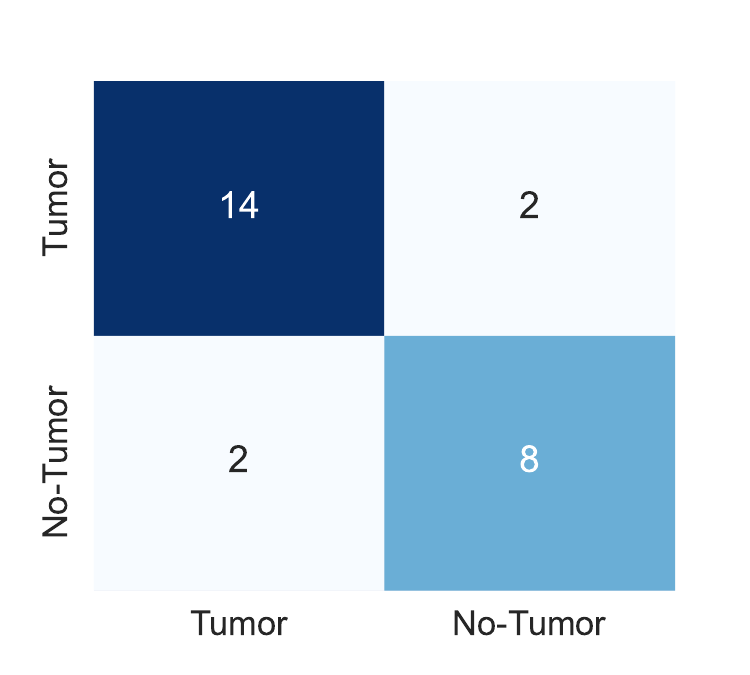}
            \subcaption{ResNet50}
            \label{CNNBr35H}
		\end{subfigure}
		\begin{subfigure}[b]{.28\columnwidth}
		    \centering
			\includegraphics[width=1\linewidth]{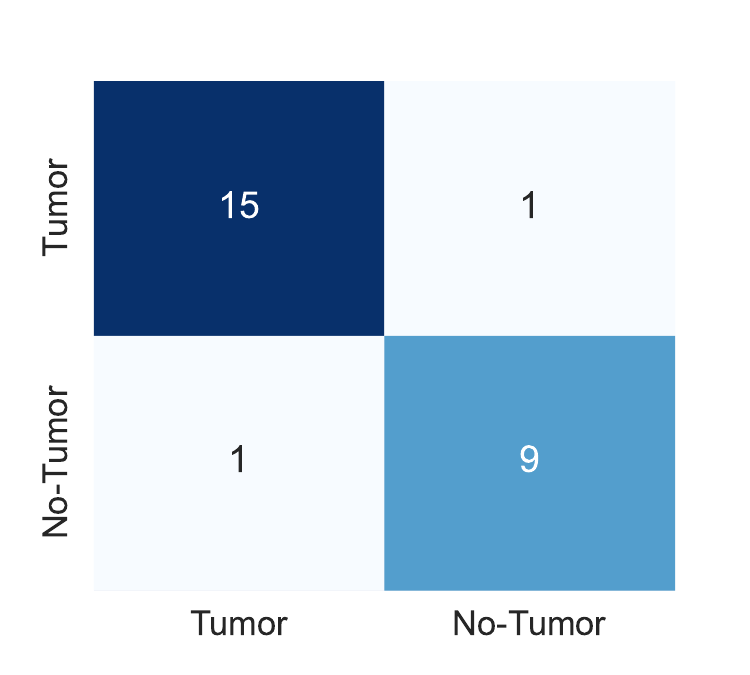}
            \subcaption{InceptionV3}
            \label{ResBTD}
		\end{subfigure}
		\begin{subfigure}[b]{.28\columnwidth}
		    \centering
			\includegraphics[width=1\linewidth]{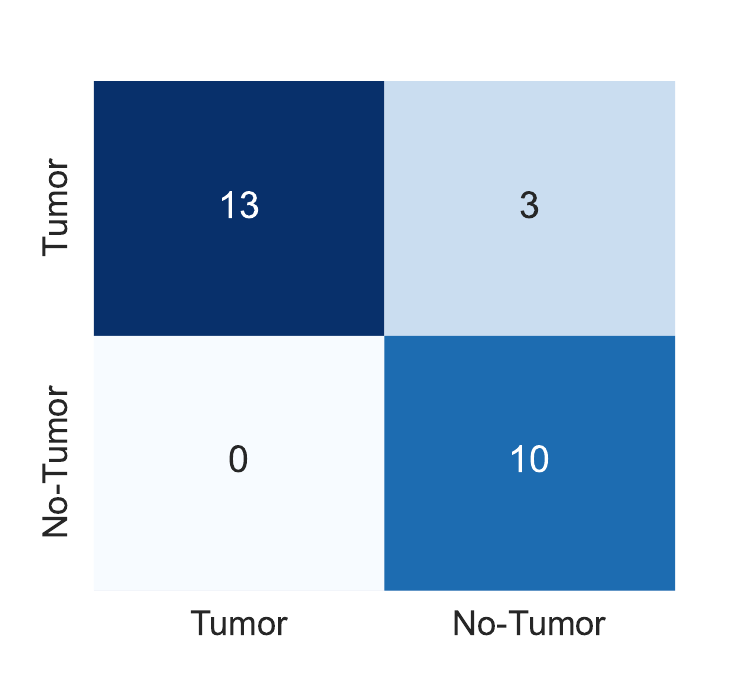}
            \subcaption{EfficientNetB0 }
            \label{CNNBTD}
		\end{subfigure}
            \begin{subfigure}[b]{.28\columnwidth}
		    \centering
			\includegraphics[width=1\linewidth]{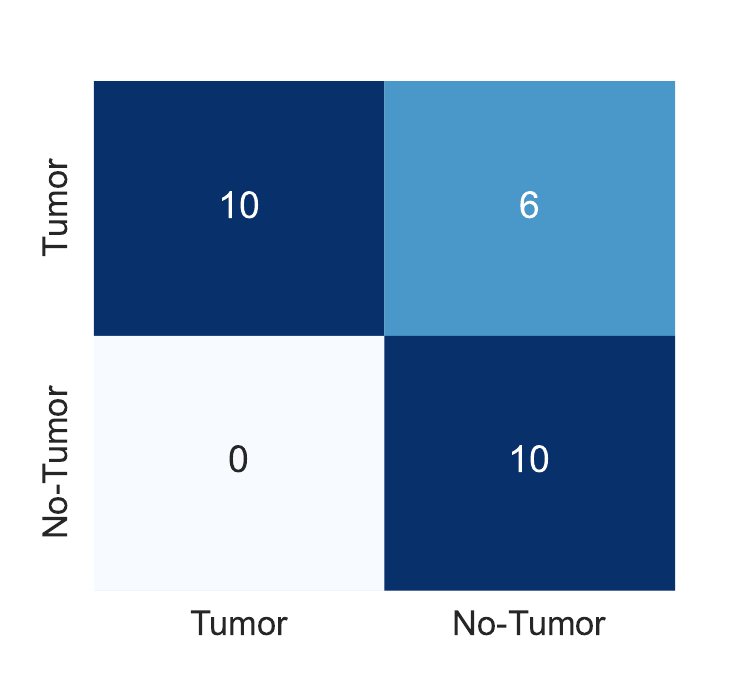}
            \subcaption{NASNetMobile}
            \label{CSResBTD}
		\end{subfigure}
            
		\caption{Confusion matrices of all the models on \textit{Brain MRI Images for Brain Tumor Detection (BTD)} dataset with equal class weight.}
		\label{fig:confusion_BTD}
	\end{figure}

\begin{figure}[h]

        \centering
		\begin{subfigure}[b]{.28\columnwidth}
		    \centering
			\includegraphics[width=1\linewidth]{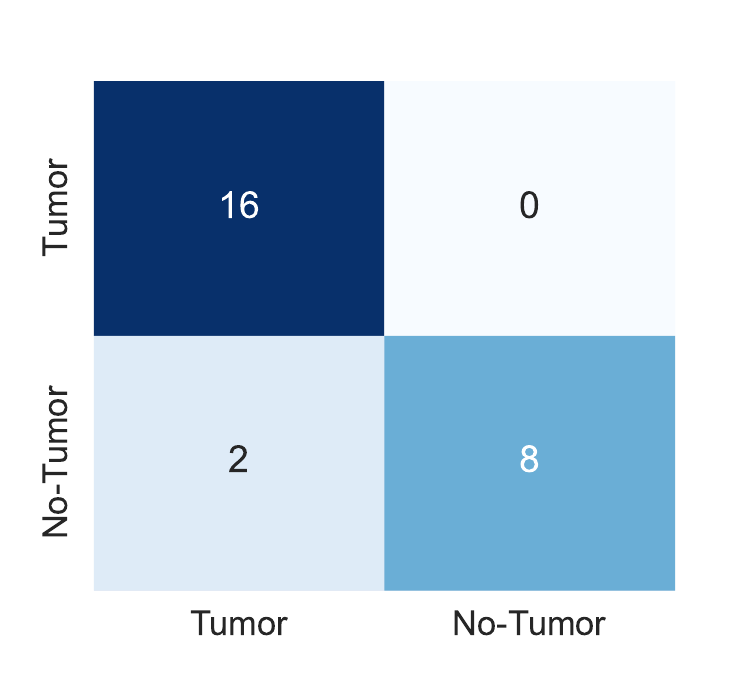}
            \subcaption{CNN}
            \label{ResBr35H}
		\end{subfigure}
		\begin{subfigure}[b]{.28\columnwidth}
		    \centering
			\includegraphics[width=1\linewidth]{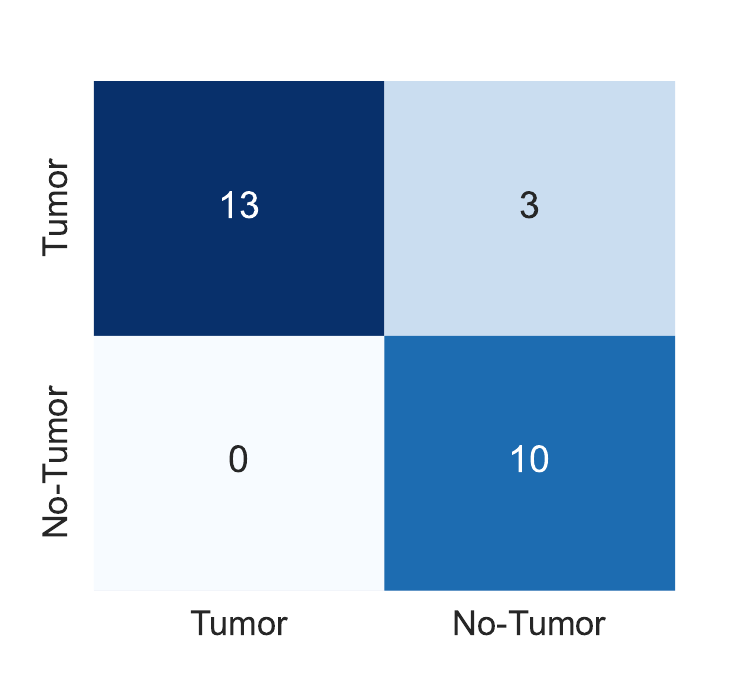}
            \subcaption{ResNet50}
            \label{CNNBr35H}
		\end{subfigure}
		\begin{subfigure}[b]{.28\columnwidth}
		    \centering
			\includegraphics[width=1\linewidth]{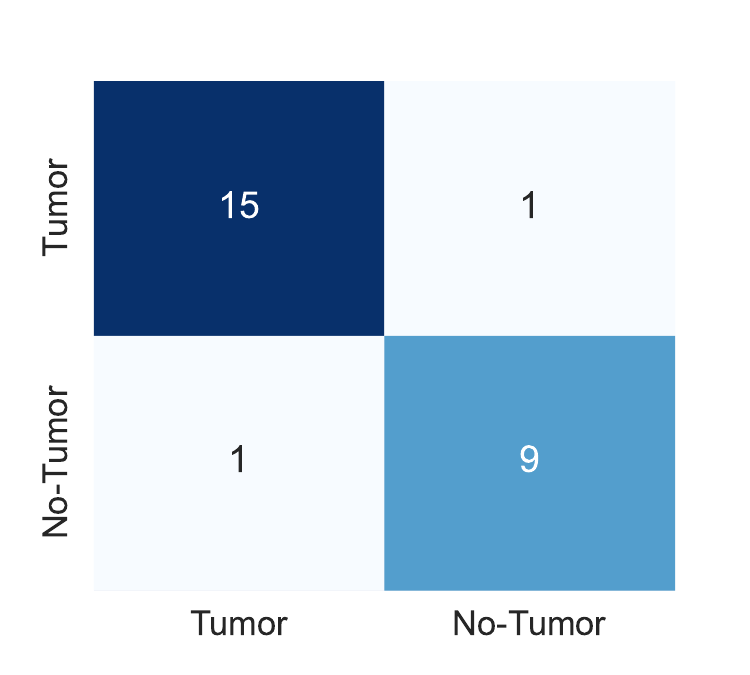}
            \subcaption{InceptionV3}
            \label{ResBTD}
		\end{subfigure}
		\begin{subfigure}[b]{.28\columnwidth}
		    \centering
			\includegraphics[width=1\linewidth]{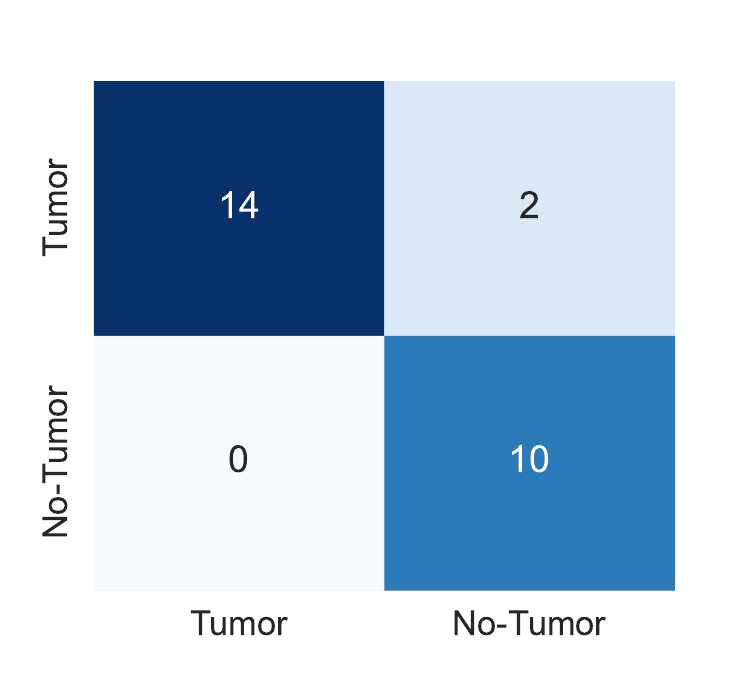}
            \subcaption{EfficientNetB0 }
            \label{CNNBTD}
		\end{subfigure}
            \begin{subfigure}[b]{.28\columnwidth}
		    \centering
			\includegraphics[width=1\linewidth]{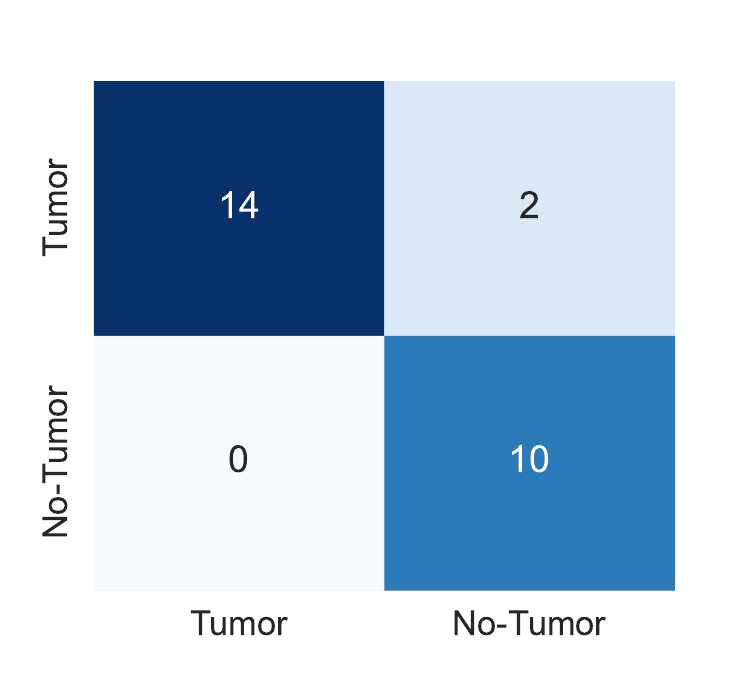}
            \subcaption{NASNetMobile}
            \label{CSResBTD}
		\end{subfigure}
            
		\caption{Confusion matrices of all the models on \textit{Brain MRI Images for Brain Tumor Detection (BTD)} dataset with cost-sensitive weight. }
		\label{fig:confusion_CSBTD}
	\end{figure}

\begin{figure}[h]

        \centering
		\begin{subfigure}[b]{.32\columnwidth}
		    \centering
			\includegraphics[width=1\linewidth]{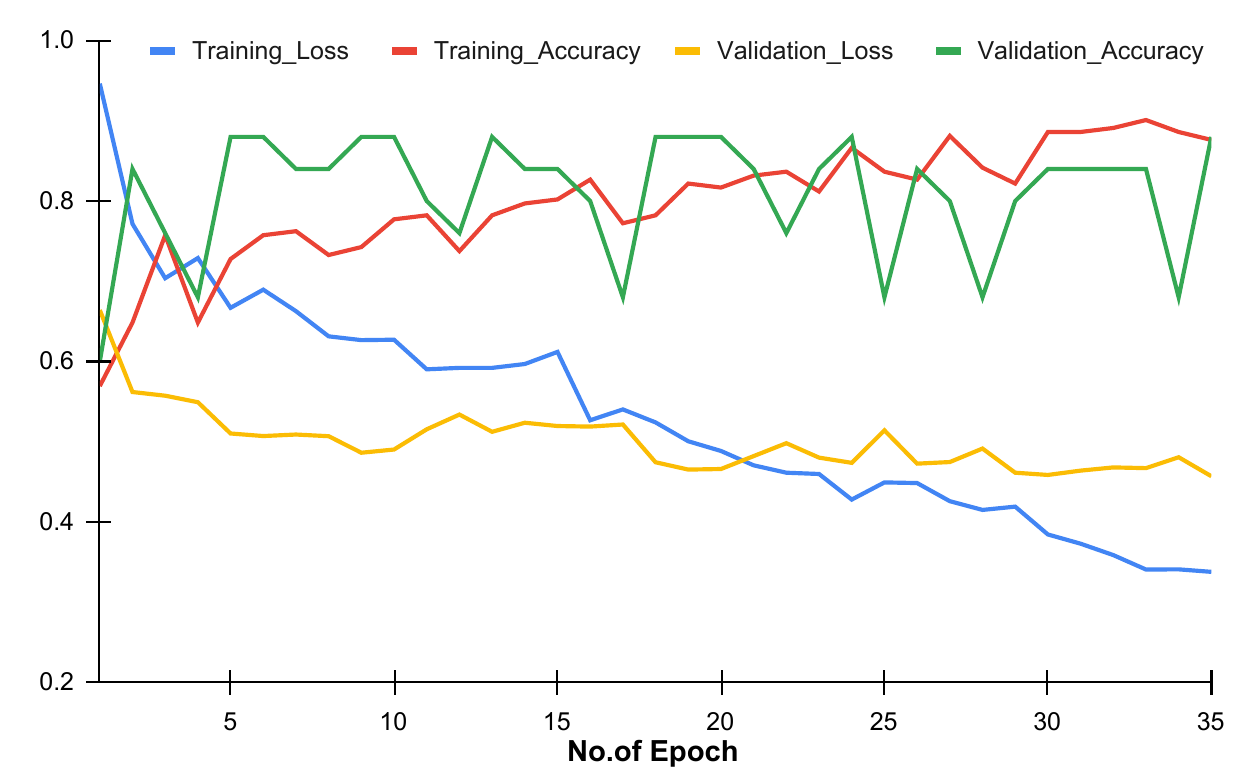}
             \subcaption{CNN}
		\end{subfigure}
		\begin{subfigure}[b]{.32\columnwidth}
		    \centering
			\includegraphics[width=1\linewidth]{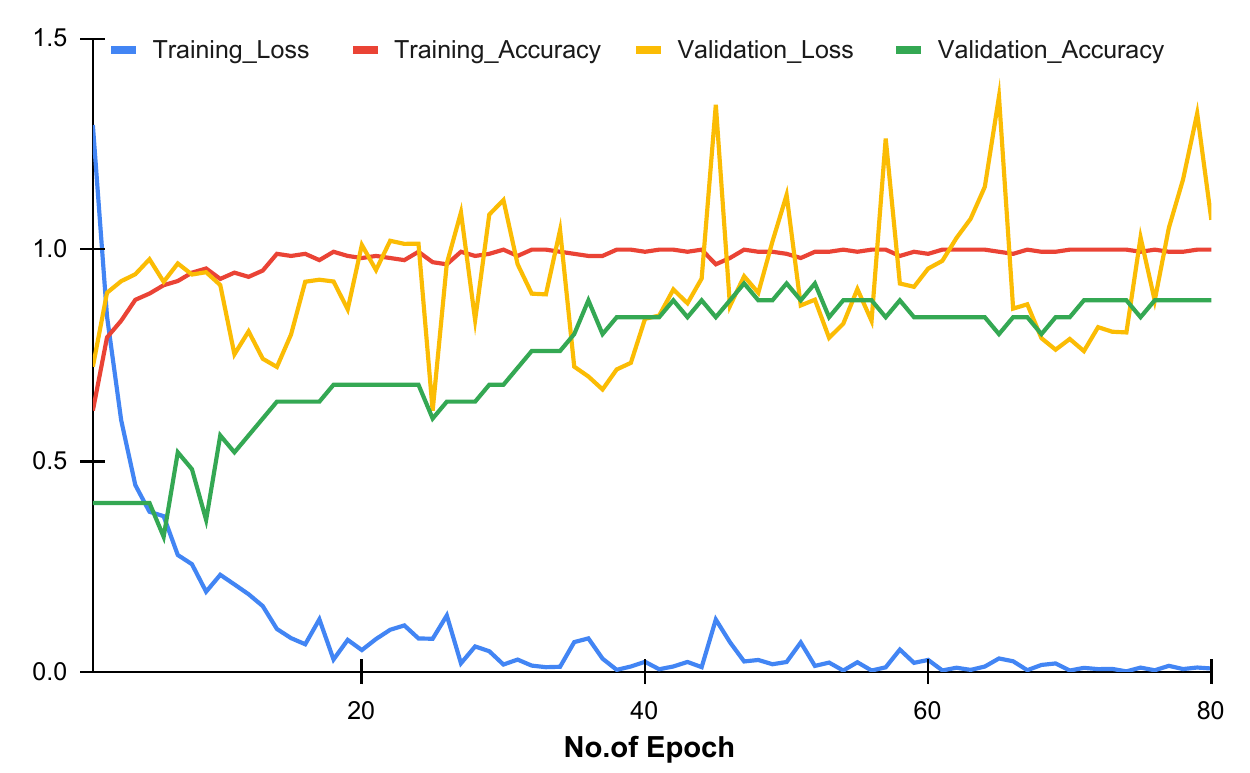}
             \subcaption{ResNet50}
		\end{subfigure}
            \begin{subfigure}[b]{.32\columnwidth}
		    \centering
			\includegraphics[width=1\linewidth]{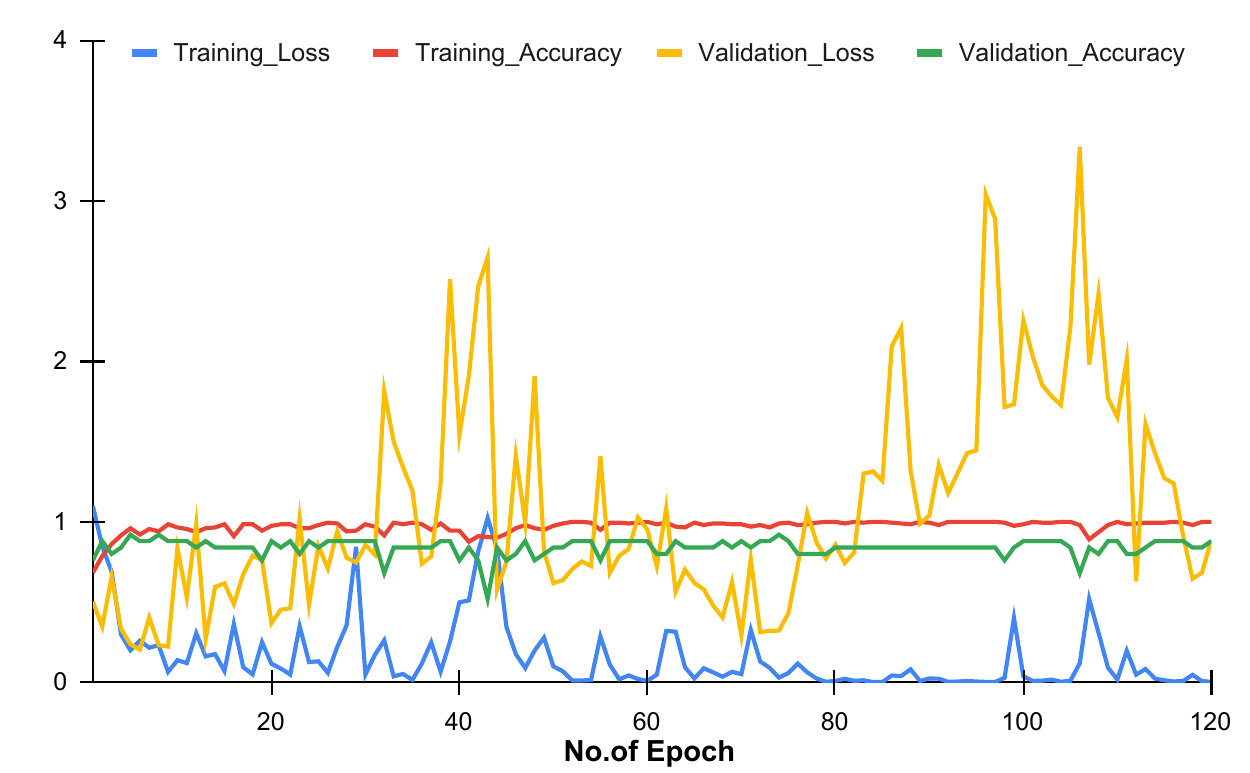}
            \subcaption{InceptionV3}
		\end{subfigure}
            \begin{subfigure}[b]{.32\columnwidth}
		    \centering
			\includegraphics[width=1\linewidth]{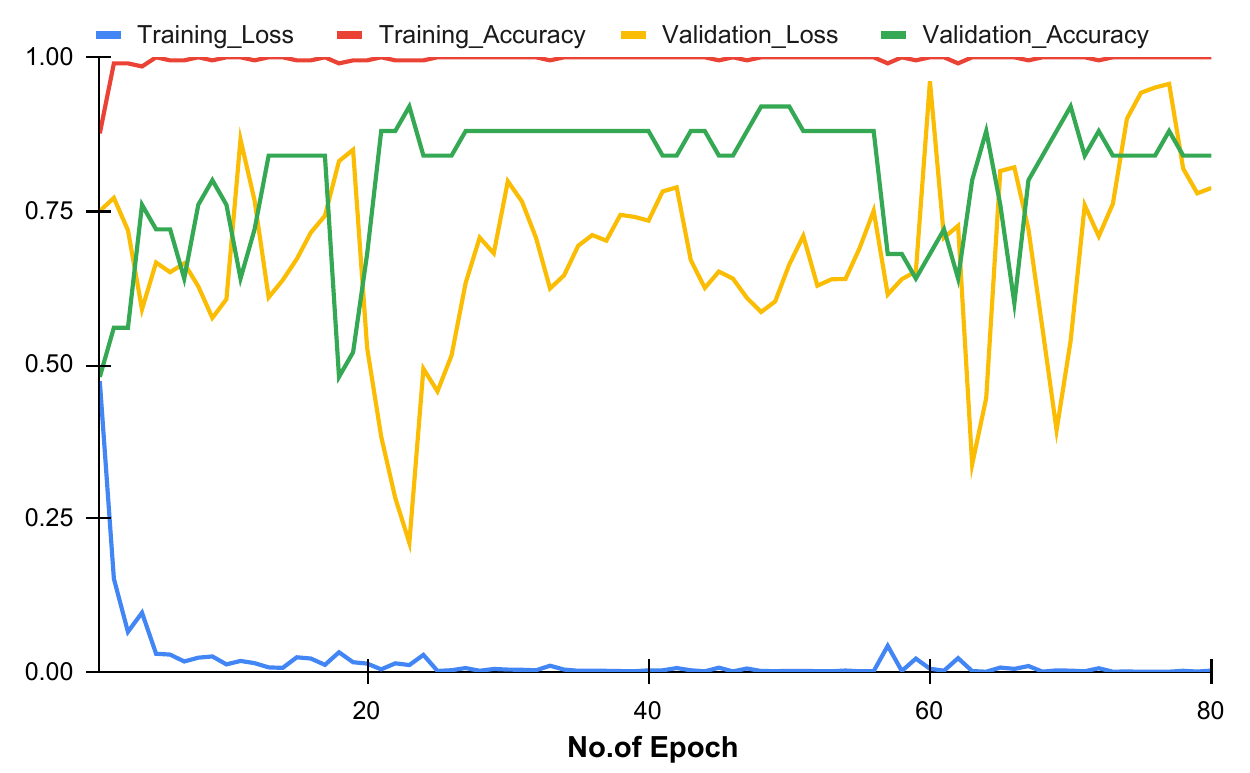}
             \subcaption{EfficientNetB0}
		\end{subfigure}
            \begin{subfigure}[b]{.32\columnwidth}
		    \centering
			\includegraphics[width=1\linewidth]{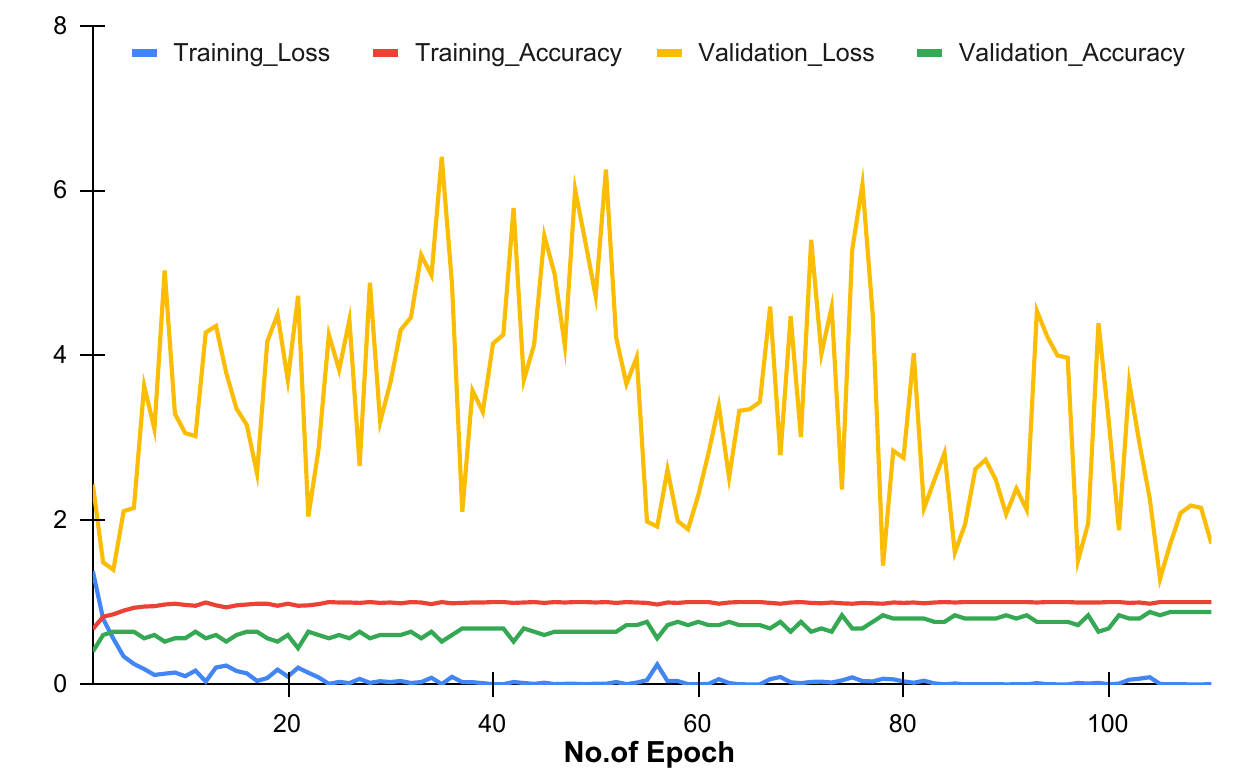}
            \subcaption{NASNetMobile}
		\end{subfigure}
		\caption{Accuracy and loss graph of training and validation for the cost-sensitive models trained on \textit{Brain MRI Images for Brain Tumor Detection} dataset.}
		\label{fig:loss2}
	\end{figure}

\begin{figure}[h]

        \centering
		\begin{subfigure}[b]{.32\columnwidth}
		    \centering
			\includegraphics[width=1\linewidth]{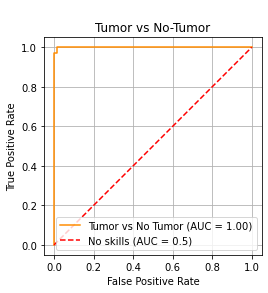}
             \subcaption{InceptionV3 (Br35H)}
		\end{subfigure}
		\begin{subfigure}[b]{.32\columnwidth}
		    \centering
			\includegraphics[width=1\linewidth]{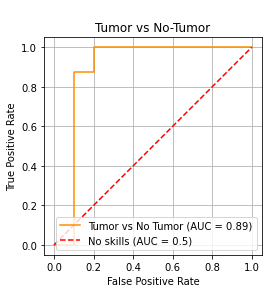}
             \subcaption{CS-CNN (BTD)}
		\end{subfigure}
            \begin{subfigure}[b]{.32\columnwidth}
		    \centering
			\includegraphics[width=1\linewidth]{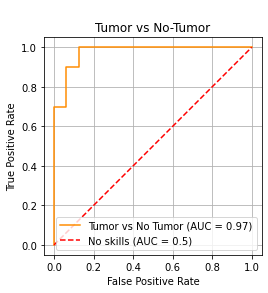}
            \subcaption{CS-InceptionV3 (BTD)}
		\end{subfigure}
           
		\caption{ROC curves for best performing models on both \textit{Br35H} and \textit{Brain MRI Images for Brain MRI Images for Brain Tumor Detection} (BTD) datasets.}
		\label{fig:roc}
	\end{figure}

			
\noindent\textbf{(c) \underline{Experimental Results}}:
 We have used Accuracy, Precision, Recall, F1-Score and Specificity as the metrics to evaluate the performance of the models. The performance of all the explored models for both dataset are shown in Table \ref{tab:accuracy}. Figure \ref{fig:confusion_Br} shows the confusion Matrices of the models on Br35H dataset. Confusion matrices of conventional and cost-sensitive models on \textit{Brain MRI Images for Brain Tumor Detection} dataset can be found in Figure \ref{fig:confusion_BTD} and \ref{fig:confusion_CSBTD} respectively. The accuracy and loss (train and validation) for cost-sensitive CNN and cost-sensitive pretrained models on \textit{Brain MRI Images for Brain Tumor Detection} dataset can be seen in Figure \ref{fig:loss2}. Figure \ref{fig:roc} shows the ROC curve of our proposed models. 

\begin{table}[]
\centering
\caption{Performance comparison among different models utilized in the proposed methodology on both the datasets. Here, CS denotes cost-sensitive.}
\label{tab:accuracy}
\resizebox{\columnwidth}{!}
{\begin{tabular}{ccccccc}
\toprule
\textbf{Dataset}                                                                                                    & \textbf{Model}                   & \textbf{Accuracy}(\%) & \textbf{Precision} & \textbf{Recall} & \textbf{F1-Score} & \textbf{Specificity}\\ \midrule
\multirow{2}{*}{Br35H}                                                      & CNN                     & 94.66       &    1  & .8933   & .9437  &  1 \\ 
& ResNet50                &    \textbf{99.33}    &  1   &  \textbf{.9867} & .9933 & 1   \\
        & InceptionV3                &    \textbf{99.33}    &  1   &  \textbf{.9867} & .9933 & 1 \\
        & EfficientNetB0       &    99.00    &  1   &  .9800 & .9899 & 1 \\
        & NASNetMobile         &   99.00     &  1   & .9800  & .9899 & 1\\
       \midrule
\multirow{10}{*}{\begin{tabular}[c]{@{}c@{}}Brain \\MRI \\ Images \\for\\ Brain \\Tumor\\ Detection \\(BTD) \end{tabular}} & CNN                     & 88.46       & .9333     & .8750  & .9032 & 0.9    \\ 
        & ResNet50                &   84.62     &  .8750   & .8750  & .8750 & 0.8    \\
        & InceptionV3                &   92.31     &  .9375   &  .9375 & .9375 & 0.9 \\
        & EfficientNetB0                &   88.46     &   1  &  .8125 & .8966  & 1 \\
        & NASNetMobile                &    76.92    &   1  & .6250  &  .7692 & 1
          \\ \cline{2-7} 
        & CS-CNN      &     92.31   &  .8888   & \textbf{1}  &  .9412 & 0.8 \\ 
        & CS-ResNet50                &    88.46    &  1   & .8125  & .8966 &  1  \\
        & CS-InceptionV3                &   \textbf{92.31}     &  .9375   & \textbf{.9375}  & .9375 & 0.9 \\
        & CS-EfficientNetB0                &    92.31    &  1   & .8750  & .9333 & 1\\
        & CS-NASNetMobile                &   92.31     &  1   & .8750  & .9333 & 1\\ \midrule
\end{tabular}}
\end{table}

			



	

\section{Result Analysis} \label{sec8}
This section is divided into three subsections. The performance of the model is covered in the first sub section, while some of the test data are analyzed with explainable AI techniques to visualize the black box of the models we used in the second subsection. The last part shows a ROC analysis of the proposed models.
\subsection{Performance of the Models}
Considering the datasets we used, the performance is also broken into two separate parts where the models trained on Br35H and Brain MRI Images for Brain Tumor Detection dataset are explained in the first and second part respectively.\\
\noindent\textbf{(a) \underline{Br35H Dataset}}:
The performance of our explored model on Br35H, which is employed as a balanced dataset in our work, is quite excellent. All of the fine-tuned pre-trained CNN models exhibit accuracy of at least 99\%. Only CNN is unable to produce a good outcome. The CNN model's narrow architecture could be to blame for its poor performance. ResNet50 and InceptionV3 provide the highest accuracy of 99.33\% with an incredible recall value of 0.9867, as can be shown in Table \ref{tab:accuracy}. Since false negative cases are more prevalent in this situation, recall is the most important factor to consider when classifying medical data, which is why we expressly highlight it in the analysis. With a good recall value of 0.98 and a specificity (true negative rate) score of 1, EfficientNetB0 and NASNetMobile also demonstrate a 99\% accuracy. The fact that all models display the same specificity score of 1 demonstrates their ability to identify negative cases. Figure \ref{fig:confusion_Br}  displays the confusion matrix for all models trained and tested on the Br35H dataset. 

\noindent\textbf{(b) \underline{Brain MRI Images for Brain Tumor Detection Dataset (BTD)}}: We performed two separate experiments on the BTD dataset based on the amount of data in each class. Because the dataset was imbalanced, we experimented with conventional CNN and four large pre-trained CNN models as well as the cost-sensitive CNN and cost-sensitive pre-trained CNN models that were previously reported. 

\begin{itemize}[wide, labelwidth=!, labelindent=0pt]
    \item \textbf{Conventional CNN and Pre-trained Models:} 
    The BTD dataset is used here to investigate the conventional CNN and four distinct pre-trained CNN models. With an accuracy of 92.31 percent and a recall value of 0.9375, InceptionV3 is the model that performed the best in this study, according to Table \ref{tab:accuracy}. The specificity is also satisfactory with a score of 0.9. CNN and EfficientNetB0 both achieve the same accuracy (88.46\%) but CNN does it with a greater recall. With the second-highest recall value of 0.8750, performance of ResNet50 is likewise pretty strong. The model with the lowest performance in the conventional approach is NASNetMobile, however, it has a specificity score of 1, indicating that it can identify negative cases but cannot predict positive cases very well. Figure \ref{fig:confusion_BTD} presents the confusion matrix of all the conventional models where we can see the dominance of InceptionV3 over all the models.

    \item \textbf{Cost-Sensitive CNN and Pre-trained CNN Models:} 
    In this part, we discuss the effectiveness of our investigated models with hard-coded class weights that preserve the proportion of classes with and without tumors. Here, we can observe that almost all of the models have outperformed conventional models in terms of accuracy. The accuracy of CNN, ResNet50, EfficientNetB0, and NASNetMobile is higher than that of the traditional ones by 3.85\%, 3.84\%, 3.85\%, and 15.39\%, respectively.The cost-sensitive CNN outperforms all other models in terms of recall(1) but falls short in terms of precision. CNN, InceptionV3, EfficientNetB0, and NASNetMobile all exhibit the same accuracy in the cost-sensitive situation, however InceptionV3 is regarded as the top model because it has the highest precision and recall value of 0.9375. But it should be mentioned that With a score of 1, ResNet50, EfficientNetB0, and NASNetMobile outperform InceptionV3 in terms of specificity. These analyses can also be verified with the confusion matrix of all the cost-sensitive models given in Figure \ref{fig:confusion_CSBTD}. The train and test loss and accuracy of all cost-sensitive models are shown in Figure \ref{fig:loss2}, where we can see how loss and accuracy fluctuate as epochs increase. These fluctuations exhibit how difficult it is to train a cost-sensitive model. 
    
\end{itemize}

\subsection{ROC Analysis}
The receiver operating characteristic curve (ROC curve) is a graph that displays how well a classification model performs across all categorization thresholds. The True Positive Rate (TPR) and False Positive Rate (FPR) are plotted on this curve where the x-axis represents the FPR and the y-axis represents the TPR. Here, we present 3 different ROC curves for our 3 proposed models: CS-CNN for Brain MRI Images for Brain Tumor Detection (BTD) dataset and InceptionV3 and cost-sensitive InceptionV3 for the Br35H and BTD datasets respectively in Figure \ref{fig:roc}. When the Area Under the Curve (AUC) score is equal to or less than 0.5, it is assumed that the model is predicting randomly and has no predictive ability. With an AUC score of 1, the ROC curve of InceptionV3 for the Br35H dataset demonstrates the model's capabilities in terms of all classification thresholds. In case of cost-sensitive CNN, ROC curve displays a good rise in y-axis as the model's TPR (Recall) is 1 but falls short in FPR. The model's overall AUC score is 0.89, which is a good outcome given that the dataset (BTD) utilized for it is an imbalanced dataset with far fewer samples than the Br35H dataset. Instead of being trained on an imbalanced dataset (BTD) our proposed cost-sensitive InceptionV3 shows an excellent performance which can be seen in the ROC curve of InceptionV3. The model's ability to accurately predict both the tumor and non-tumor is demonstrated by the AUC score of 0.97.

\subsection{Explainable AI}
When presenting the CNN based model, we took into account Vanilla Saliency, SmoothGrad, Grad-CAM, Grad-CAM++, Score-CAM, and Faster Score-CAM, while LIME is used to explain the ResNet50, InceptionV3, EfficientNetB0 and NASNetMobile models. Because of LIME's differences from other techniques, LIME has been separately detailed in this subsection.
\begin{figure}[t]
		\centering
		\begin{subfigure}[b]{1\columnwidth}
		    \centering
			\includegraphics[width=1\linewidth]{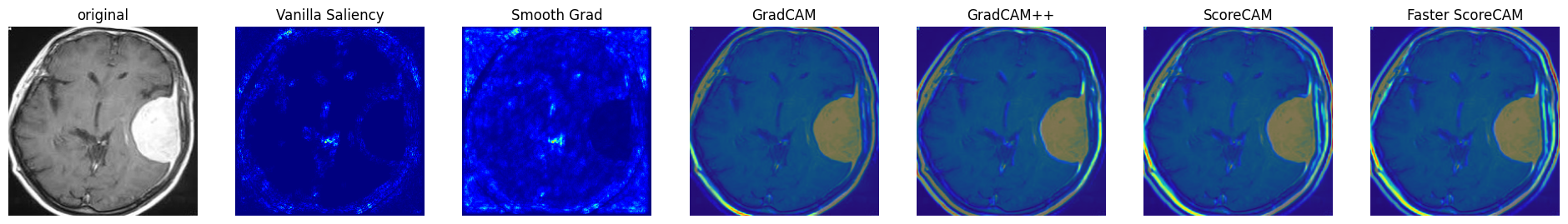}
		\end{subfigure}
		\begin{subfigure}[b]{1\columnwidth}
		    \centering
			\includegraphics[width=1\linewidth]{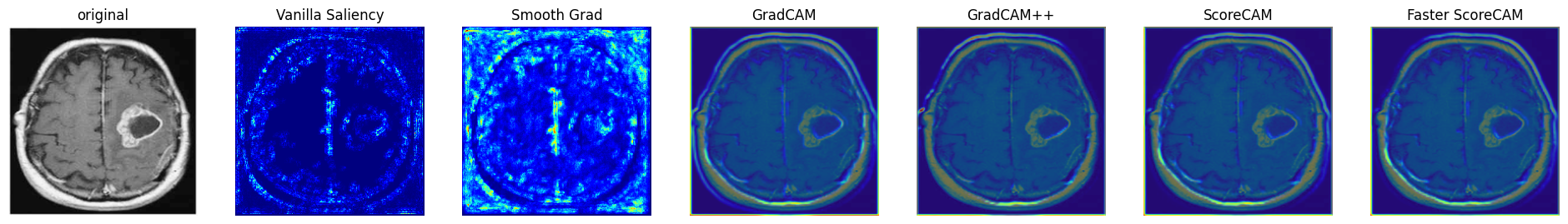}
		\end{subfigure}

		\caption{Visualization of sample outputs of two correctly classified (\textit{top}) and miss-classified (\textit{bottom}) instances of \textbf{Tumor} category on \textit{Br35H} dataset using CNN. \textbf{($1^{st}$ row} - True Label:1, Predicted:1 and \textbf{$2^{nd}$ row} - True Label:1, Predicted:0 )}

		\label{fig:Br35HGrad}
	\end{figure}

\begin{figure}[t]
		\centering
		\begin{subfigure}[b]{1\columnwidth}
		    \centering
			\includegraphics[width=1\linewidth]{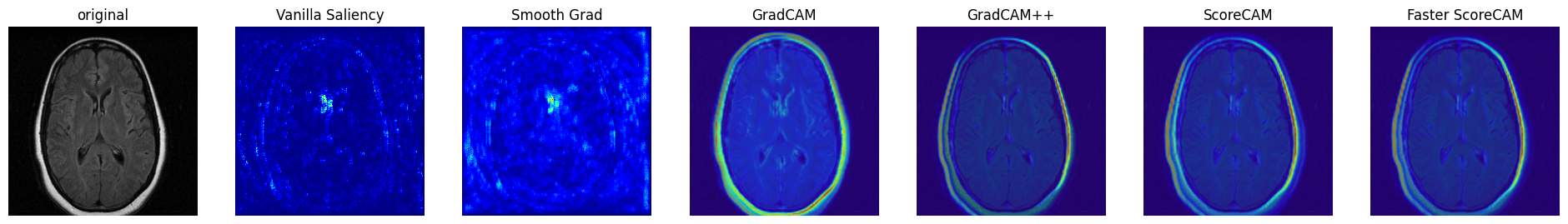}
		\end{subfigure}
		\begin{subfigure}[b]{1\columnwidth}
		    \centering
			\includegraphics[width=1\linewidth]{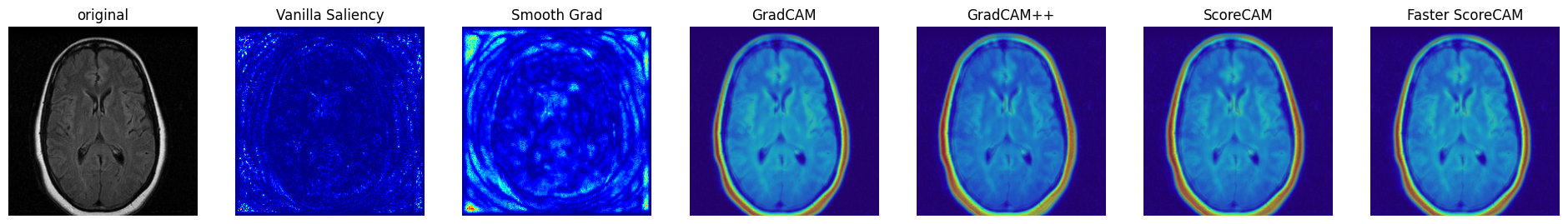}
		\end{subfigure}

		\caption{Visualization of sample outputs of a misclassified instance of \textbf{No-Tumor} category on \textit{Brain MRI Images for Brain Tumor Detection} dataset using CNN (top) and CS-CNN (bottom). Both the models give wrong prediction.}
		\label{fig:DetectionGrad}
	\end{figure}
 
 \begin{figure}[t]
		\centering
		\begin{subfigure}[b]{1\columnwidth}
		    \centering
			\includegraphics[width=1\linewidth]{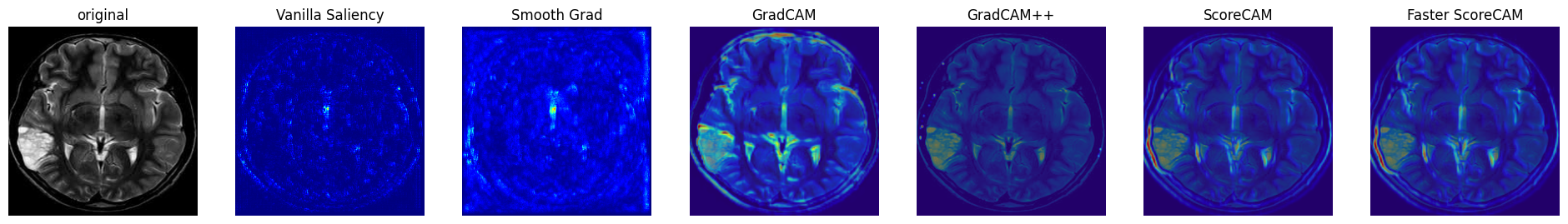}
		\end{subfigure}
		\begin{subfigure}[b]{1\columnwidth}
		    \centering
			\includegraphics[width=1\linewidth]{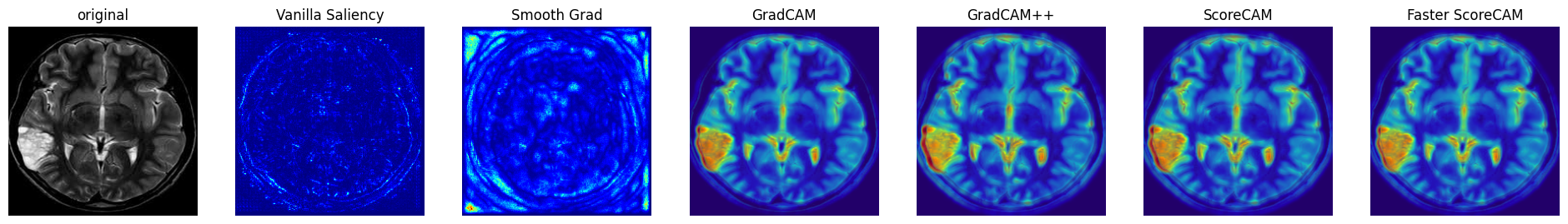}
		\end{subfigure}
		\caption{Visualization of the outputs of a sample instance of \textbf{Tumor} category on \textit{Brain MRI Images for Brain Tumor Detection} dataset by CNN (top) and CS-CNN (bottom) where cost-sensitive CNN correctly classifies the instance but standard CNN misclassifies.}
		\label{fig:DetectionGradmiss}
	\end{figure}
\subsubsection*{(a) \underline{Gradient-based Explainable AI methods}}
\begin{itemize}[wide, labelwidth=!, labelindent=0pt]
\item \textbf{Br35H Dataset}: The Convolutional Neural Network model, which has been shown to be one of the best classification models, is a strong model for extracting features from images. In this task, CNN has been explained with the use of different explainable AI approaches. Figure \ref{fig:Br35HGrad} depicts two accurately identified test samples, one of which is appropriately identified as a tumor and another as a non-tumor. The top row shows a correctly classified sample. The original image is on the left, and each of the other explainable AI outputs is shown side by side. Vanilla saliency and SmoothGrad show the area of interests, but it is not sufficiently enough to explain the model's behavior. Grad-CAM has good efficiency in highlighting the focal points of the image, and it is obvious that the model has the ability to identify the region of interest. Grad-CAM++, Score-CAM and Faster Score-CAM provides almost similar realistic explanation as the Grad-CAM, demonstrating clear region of interests and the outer boundary of the brain image. The bottom image shows a misclassification example where the original image contains a tumor but the CNN model predicted wrong and could not identify the tumor region which is evident from the outputs of the explainable AI approaches. 
    
\item \textbf{Brain MRI Images for Brain Tumor Detection (BTD) Dataset}:
We perform a normal CNN as well as a cost-sensitive CNN while taking into account the data equivalence in each class. We have examined both models using various explainable AI approaches. 
\begin{table}[]
\centering
\caption{Prediction probability of \textbf{Tumor} and \textbf{No-Tumor} class by CNN and cost-sensitive CNN (CS-CNN) for the instances in Figure \ref{fig:DetectionGrad} and \ref{fig:DetectionGradmiss}.}
\label{tab:probtable}
\begin{tabular}{cccc}
\toprule
\textbf{Figure}         & \textbf{Model} & \begin{tabular}[c]{@{}c@{}}\textbf{Probability of}\\ \textbf{Tumor Class}\end{tabular} & \begin{tabular}[c]{@{}c@{}}\textbf{Probability of}\\ \textbf{No-Tumor Class}\end{tabular} \\
\midrule
\multirow{2}{*}{\ref{fig:DetectionGrad}} & CNN 
 (\textit{top})    &  0.9999                                                                & 0.0001                                                              \\ 
                  & CS-CNN (\textit{bottom})     & 0.9871                                                                  & 0.0129                                                           \\ \midrule
\multirow{2}{*}{\ref{fig:DetectionGradmiss}} & CNN (\textit{top})  
& 0.0040                                                                  & 0.9960  \\ 
                  & CS-CNN (\textit{bottom}) & 0.8203                                                                  & 0.1797  \\      \bottomrule                                                 
\end{tabular}
\end{table}
Figure \ref{fig:DetectionGrad} illustrates an example of misclassification by both the CNN and cost-sensitive CNN models. The original image on the leftmost side does not have any tumor. The top row image presents the output explanation of the CNN model and the bottom row image presents the output explanation of the cost-sensitive CNN model. It is evident from the vanilla saliency map and SmoothGrad on the top row image that CNN model highlights a circular region (white colored) which might be considered as a tumor. Though the circular region is not clearly visible in other XAI approaches, CNN highlights the boundary region. The white colored circular region is not present in case of cost-sensitive CNN model. But due to weight adjustments, the boundary region got more attention. As a result the model might got confused which is supported by the SmoothGrad visualization where several regions are highlighted in the whole image. This indicates that features that were not important influenced the model towards incorrect prediction.
Figure \ref{fig:DetectionGradmiss} shows a sample that normal CNN incorrectly classifies while cost-sensitive CNN classifies correctly. Table \ref{tab:probtable} displays the prediction made by both the CNN and cost-sensitive CNN models for the same sample. Though the standard CNN model learns the Tumor class better because the dataset has more data in the Tumor category, it is evident from the image on the top row that the CNN model could not identify the tumor. It also failed to identify the boundary region. Though the visualizations provided by the explainable approaches highlights the region of interest (tumor) but gradually with each approach the region of interest faded. This clearly proves that the CNN model could not provide enough attention or weights to the region of interest which might be the possible cause for the misclassification.  On the other hand, from the visualizations of the explainable AI approaches on the bottom row image, we can see the power of weight biasing which can be achieved by the cost-sensitive CNN model. The visualization by the explainable AI techniques shows that the cost-sensitive CNN not only takes into account the entire brain structure but also assign more and clear weights on the region of interest for the prediction. With each explainable AI approach the region of interest gets strong contrast.  
\end{itemize}

\subsubsection*{(b) \underline{Perturbation-based Explainable AI method}}
\begin{figure}[h]
		\centering
			\includegraphics[width=0.9\linewidth]{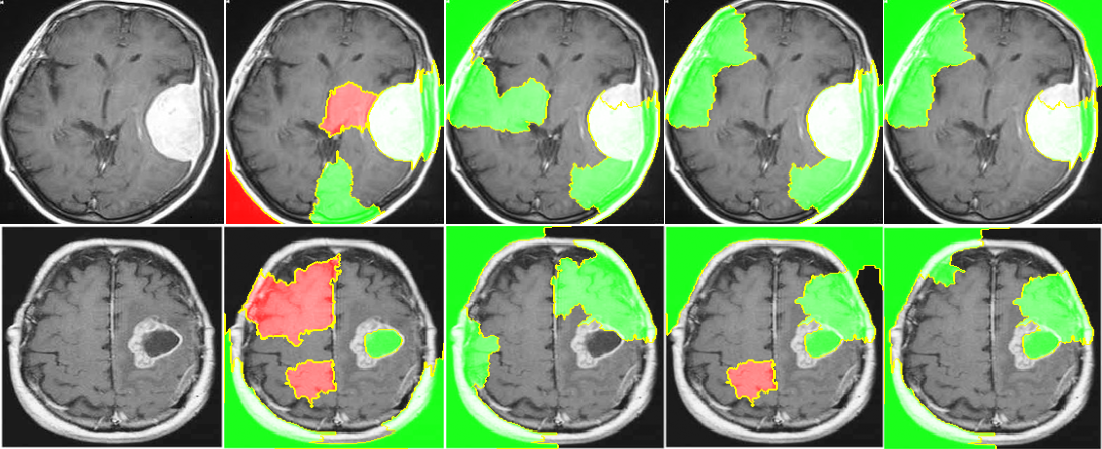}
		\caption{LIME output ($1^{st}$ and $2^{nd}$ row) of \textit{ResNet50}, \textit{InceptionV3}, \textit{EfficientNetB0}, and \textit{NASNetMobile} for a sample instance with the original image (from \textit{left} to \textit{right}) on \textbf{Br35H} dataset. (\textbf{$1^{st}$ row} - \textit{True Label}: 1, \textit{ResNet50}:0, \textit{InceptionV3}:1, 
 \textit{EfficientNetB0}:1, and 
 \textit{NASNetMobile}:1. \textbf{$2^{nd}$ row} - \textit{True Label}: 1, \textit{ResNet50}:1, \textit{InceptionV3}:0, 
 \textit{EfficientNetB0}:1, and 
 \textit{NASNetMobile}:0). Here 0 and 1 denotes \textbf{No-Tumor} and \textbf{Tumor} class respectively.}
		\label{fig:limeBr35H}
	\end{figure}
\begin{figure}[t!]
\centering

		\begin{subfigure}[b]{1\columnwidth}
		  \centering
			\includegraphics[width=0.9\linewidth]{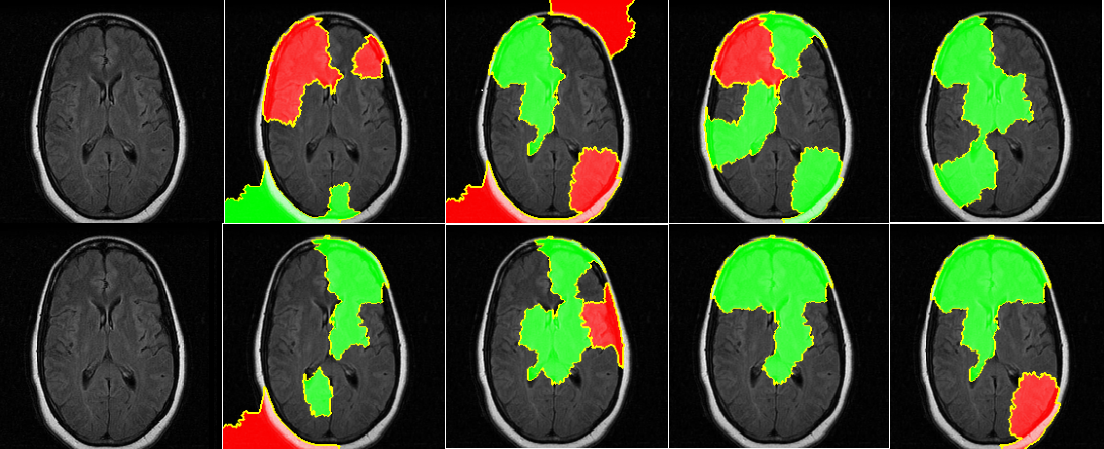}
			\subcaption{\textit{True Label}: 0, \textit{ResNet50}:1, 
 \textit{CS-ResNet50}:0, \textit{InceptionV3}:0,
    \textit{CS-InceptionV3}:0, 
 \textit{EfficientNetB0}:0,
    \textit{CS-EfficientNetB0}:0, 
 \textit{NASNetMobile}:0, and
    \textit{CS-NASNetMobile}:0}
        \label{fig:limedetectiondata_a}
		\end{subfigure}
		\begin{subfigure}[b]{1\columnwidth}
		  \centering
    \includegraphics[width=0.9\linewidth]{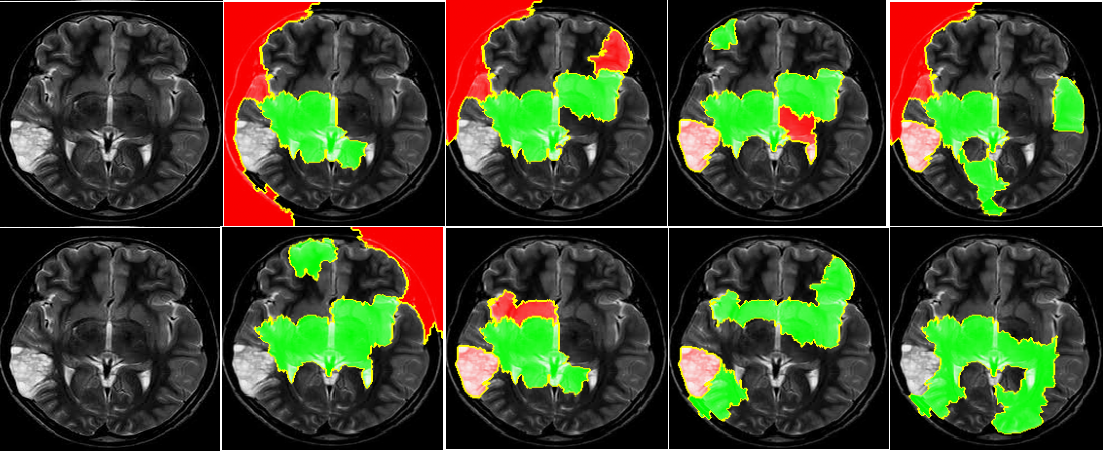}
			\subcaption{\textit{True Label}: 1, \textit{ResNet50}:0, 
 \textit{CS-ResNet50}:0, \textit{InceptionV3}:0,
    \textit{CS-InceptionV3}:0, 
 \textit{EfficientNetB0}:0,
    \textit{CS-EfficientNetB0}:0, 
 \textit{NASNetMobile}:0, and
    \textit{CS-NASNetMobile}:0}
        \label{fig:limedetectiondata_b}
		\end{subfigure}
		\caption{LIME output of \textit{ResNet50}, \textit{InceptionV3}, \textit{EfficientNetB0}, and \textit{NASNetMobile} for a sample instance with the original image (from \textit{left} to \textit{right}) with the output of their corresponding  cost-sensitive model below on \textbf{Brain MRI Images for Brain Tumor Detection} dataset. (CS denotes cost-sensitive and 0 and 1 denotes \textbf{No-Tumor} and \textbf{Tumor} class respectively.)}
		\label{fig:limedetectiondata}
	\end{figure}
\begin{itemize}[wide, labelwidth=!, labelindent=0pt]
    \item \textbf{Br35H Dataset}:
    The pretrained models that are ResNet50, InceptionV3, EfficientNetB0 and NASNetMobile gave promising results for our classification task and have proven to be efficient feature extraction models for images. In our work LIME has been used to explain these models behavior. Figure \ref{fig:limeBr35H} illustrates 10 samples showed in two different rows, for brain tumour classification, each of the sample image has tumor in it. Within each row, the original image is shown on the left, and LIME outputs for the interpretation of the ResNet50, InceptionV3, EfficientNetB0, NASNetMobile models are shown from left to right. Predicted labels of each models are also mentioned in the figure caption. In the LIME outputs, the highlighted regions are the ones which are used by the model to make a decision, of which the green marked regions are the ones that the model actively used for the correct classification whereas the red marked regions represents opposite class probability. The sample of the top row is only misclassified by ResNet50. The second image from the left on the top row depicts that the portion of tumor used by ResNet50 model to make prediction was not sufficient enough to enable it to classify it as a tumor contained image. The remaining three images were correctly classified by the rest of the models as the portion of the images focused by the models were accurate and sufficient enough to enable the models to classify the images as brain tumor contained image. In the third image of the second row, we can see that only Inception V3 wrongly classified the image as the model could not focus on the brain region while classifying the image. 

    \item \textbf{Brain MRI Images for Brain Tumor Detection Dataset}:
    The Brain MRI Images for Brain Tumor Detection dataset was an imbalanced dataset so to solve this issue, beside regular architecture, we have designed cost-sensitive Neural Network Architecture to reduce training biasness. This cost-sensitive Neural Network Architecture gives us a noticeable better performance compared to the regular Neural Network architecture. It can also be observed from the visualization of the Explainable AI outputs that these cost-sensitive architecture is comparatively effective.  Figure \ref{fig:limedetectiondata_a}  illustrates total eight cases for brain tumour classification, for each pair the right image is for the cost-sensitive model and and the left image is for the regular model. The true label for this image is 0, for which only regular ResNet50 predicted wrongly. From the LIME output illustrated on the second image from the left in the first row, it can be seen that the model focuses on the outside of the brain region to make it's decision which lead to wrong prediction. Figure \ref{fig:limedetectiondata_b} illustrates a image whose true label is 1 and all the models failed to correctly classify this image. The main problem in this case is the image itself, the shape, pattern and color distribution of the image is too complex here which does not match with the other training data.
\end{itemize}
\subsection{Performance Comparison}
In Table \ref{tab:comparison}, we present a comparison of our proposed models with other existing works done on Br35H and Brain MRI Images for Brain Tumor Detection (BTD) datasets. From Table \ref{tab:comparison}, we can observe that the study by Mondol et al. \cite{mondal2022novel} employed a deep CNN model and achieved an accuracy of 99.00\%, while Islam et al. \cite{10099302} shows an accuracy of 99.00\% with MobileNet for the Br35H dataset, where our proposed InceptionV3 obtains an accuracy of 99.33\%. With a train, validation, and test ratio of 80, 20 and 20, respectively, Sailunaz et al.\cite{sailunaz2023brain} utilized a CNN model with MRI features on the imbalanced BTD dataset and obtained an accuracy of 76.47\%. With the same ratio the authors used in their study, we tested our proposed cost-sensitive CNN, and after training the model for 30 epochs, we were able to reach an accuracy of 90.20\% with a training accuracy of 92.76\% and validation accuracy of 86.0\%. Our cost-sensitive CNN achieves a recall value of 1, whereas the study \cite{rai2019hybrid} by Sruti et al. explored a hybrid convolutional SVM on the identical imbalanced dataset and achieved a recall value of 0.98.

\begin{table}[]
\centering
\caption{Performance comparison with other existing works maintaining the same train-validation-test split.}
\label{tab:comparison}
\resizebox{\columnwidth}{!}
{\begin{tabular}{cccccc}
\hline
\textbf{Dataset}                                                                                                      & \textbf{Reference}           & \textbf{Model}                                                                 & \textbf{Metric}   & \textbf{Score} & \textbf{Our approach}                                                                    \\ \hline
\multirow{2}{*}{Br35H}                                                                                       & Mondal et al.\cite{mondal2022novel}   & Deep CNN                                                              & Accuracy & 99.00  & \multirow{2}{*}{\begin{tabular}[c]{@{}c@{}}99.33 \\ (InceptionV3)\end{tabular}} \\ \cline{2-5}& Islam et al.\cite{10099302}    & MobileNet                                                             & Accuracy & 99.00  &  \\ \hline
\multirow{2}{*}{\begin{tabular}[c]{@{}c@{}}Brain MRI \\ Images for \\ Brain Tumor \\ Detection\end{tabular}} & Sailunaz et al.\cite{sailunaz2023brain} & \begin{tabular}[c]{@{}c@{}}CNN \\ (MRI Feature)\end{tabular}          & Accuracy & 76.47  & \begin{tabular}[c]{@{}c@{}}90.20\\ (Cost-sensitive\\ CNN)\end{tabular}          \\ \cline{2-6} 
 & Sruti et al.\cite{rai2019hybrid}    & \begin{tabular}[c]{@{}c@{}}Convolutional \\ SVM (Hybrid)\end{tabular} & Recall   & 0.98   & \begin{tabular}[c]{@{}c@{}}1\\ (Cost-sensitive\\ CNN)\end{tabular}              \\ \hline
\end{tabular}}
\end{table}

\section{Limitations and Future Work}\label{sec9}
    On MRI imaging of brain tumors, there exist numerous datasets. We could only deal with two datasets at a time and could not use many hyper-parameters. Additionally, the Explainable AI approach we have utilized here only supports gradient-based models like neural networks. Because of this, we were unable to apply this XAI techniques to other machine learning techniques. In future, we can investigate several new dimensions such as: (1) finding a more accurate model for brain tumor detection task. Other pre-trained models like DenseNet, Xception, ResNext and Wide-ResNet can be explored, (2) working with other datasets created on brain tumor MRI images and experimenting with various hyper-parameters and (3) visualizing with new Explainable AI techniques as this area of research is growing exponentially. 

\section{Conclusion} \label{sec10}
This research provides promising results for brain tumor detection using CNN and pre-trained models. The fine-tuned InceptionV3 achieved an excellent accuracy of 99.33\% for a balanced dataset. Moreover, the cost-sensitive use of these models provided greater accuracy than standard ones in an imbalanced dataset. The cost-sensitive InceptionV3 (CS-InceptionV3) and CNN (CS-CNN) show a promising accuracy of 92.31\% and a recall value of 1 respectively. However, the application of explainable AI technique provides a better understanding of features and the fundamentals behind its decision-making process, thus helping to understand more about the black box component involved in this machine learning process. These findings suggest that cost-sensitive CNN (CS-CNN) and InceptionV3 (CS-InceptionV3) are effective for brain tumor detection and can be used in diagnostic applications. Finally, We believe this research can help the researchers understand the inner mechanism of neural networks and assist people to get an earlier alert about their health state. However, more study is required to boost the accuracy of the models and make them suitable for use in the actual world.


\section{Declarations}
\subsection{Ethical Approval and Consent to participate}
Not Applicable.
\subsection{Consent for publication}
Not Applicable.
\subsection{Human and Animal Ethics}
Not Applicable.
\subsection{Competing interests}
The authors declare that they have no competing interests.
\subsection{Funding}
No funding was received for conducting this study.
\subsection{Authors' contributions}
All authors contributed equally to this work. 
\subsection{Data and Code availability}
The train, validation and test split of the data used in this study and all the implementations for the experimentation purpose are publicly available at -  \url{https://github.com/shahariar-shibli/Explainable-Cost-Sensitive-Deep-Neural-Networks-for-Brain-Tumor-Detection-from-Brain-MRI-Images}\\
The balanced dataset \textit{Br35H :: Brain Tumor Detection 2020} can be found at - \url{https://www.kaggle.com/datasets/ahmedhamada0/brain-tumor-detection}\\
The imbalanced dataset \textit{Brain MRI Images for Brain Tumor Detection} can be found at -\url{https://www.kaggle.com/datasets/navoneel/brain-mri-images-for-brain-tumor-detection}

\bibliography{sn-bibliography}


\end{document}